\pdfoutput=1

\documentclass[11pt]{article}

\usepackage{acl}

\usepackage{times}
\usepackage{latexsym}

\usepackage[T1]{fontenc}

\usepackage[utf8]{inputenc}

\usepackage{microtype}

\usepackage{amsthm}
\usepackage{amsmath}
\usepackage{amssymb}
\usepackage{booktabs}
\usepackage{graphicx}
\usepackage{subcaption}
\usepackage{multirow}
\usepackage{xcolor}

\newtheorem{lemma}{Lemma}
\newtheorem{conjecture}{Conjecture}

\colorlet{40Red}{red!40}
\colorlet{45Red}{red!45}
\colorlet{50Red}{red!50}
\colorlet{70Red}{red!70}
\colorlet{70BGreen}{green!70!black}

\title{Towards Better Understanding of Contrastive Sentence Representation Learning: A Unified Paradigm for Gradient}

\author{
    Mingxin Li\textsuperscript{\rm 1}, Richong Zhang\textsuperscript{\rm 1,2}\thanks{\ \ Corresponding author}, Zhijie Nie\textsuperscript{\rm 1,3}\\
    \textsuperscript{\rm 1}CCSE, School of Computer Science and Engineering, Beihang University, Beijing, China\\
    \textsuperscript{\rm 2}Zhongguancun Laboratory, Beijing, China\\
    \textsuperscript{\rm 3}Shen Yuan Honors College, Beihang University, Beijing, China\\
    \texttt{\{limx,zhangrc,niezj\}@act.buaa.edu.cn}
}

\begin{document}
\maketitle
\begin{abstract}
Sentence Representation Learning (SRL) is a crucial task in Natural Language Processing (NLP), where contrastive Self-Supervised Learning (SSL) is currently a mainstream approach. However, the reasons behind its remarkable effectiveness remain unclear. 
Specifically, many studies have investigated the similarities between contrastive and non-contrastive SSL 
from a theoretical perspective. Such similarities can be verified in classification tasks, where the two approaches achieve comparable performance. But in ranking tasks (i.e., Semantic Textual Similarity (STS) in SRL), contrastive SSL significantly outperforms non-contrastive SSL.
Therefore, two questions arise: First, \textit{what commonalities enable various contrastive losses to achieve superior performance in STS?} Second, 
\textit{how can we make non-contrastive SSL also effective in STS?}
To address these questions, we start from the perspective of gradients and discover that four effective contrastive losses can be integrated into a unified paradigm, which depends on three components: the \textbf{Gradient Dissipation}, the \textbf{Weight}, and the \textbf{Ratio}. Then, we conduct an in-depth analysis of the roles these components play in optimization and experimentally demonstrate their significance for model performance. Finally, by adjusting these components, we enable non-contrastive SSL to achieve outstanding performance in STS. 
\footnote{Our code is available at \url{https://github.com/BDBC-KG-NLP/UnderstandingCSE}.}
\end{abstract}


\begin{figure}[th]
    \centering
    \includegraphics[width=0.48\textwidth]{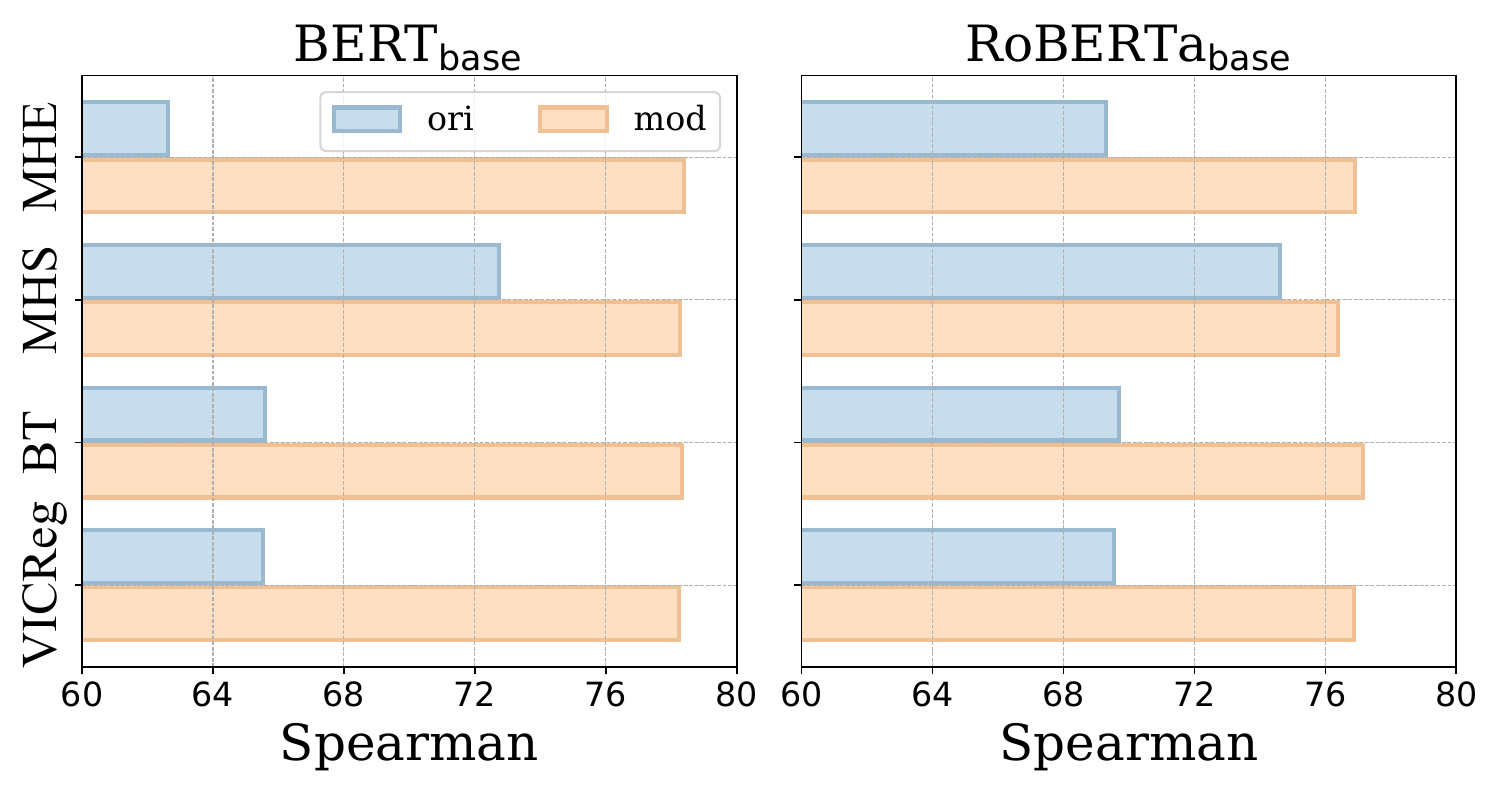}
    \caption{Average Spearman's correlation on Semantic Textual Similarity tasks for ineffective optimization objectives before (``ori'') and after (``mod'') modifications under different backbones.}
    \label{fig:performance_0}
\end{figure}

\section{Introduction}
\label{sec:introduction}

Sentence Representation Learning (SRL) explores how to transform sentences into vectors (or ``embeddings''), which contain rich semantic information and are crucial to many downstream tasks in Natural Language Processing (NLP). In the era of Large Language Models (LLMs), SRL also plays an important role in providing the embedding models for Retrieval-Augmented Generation (RAG, \citealp{gao_retrieval-augmented_2023}). 
The quality of sentence embeddings is usually measured through Transfer tasks (TR) and Semantic Textual Similarity tasks (STS)~\citep{conneau_senteval_2018}.

Contrastive Self-Supervised Learning (SSL) is now a prevalent approach in SRL, which is introduced by \citet{gao_simcse_2021} and \citet{yan_consert_2021}. It optimizes the representation space by reducing the distance between a sentence (or ``anchor'') and semantically similar sentences (or ``positive samples''), 
as well as increasing the distance between the sentence and semantically dissimilar sentences (or ``negative samples''). 

While the mechanisms underlying contrastive SSL can be intuitively understood, its effectiveness in SRL has not been thoroughly explored. Specifically, there are still some conflicts between the existing conclusions:
(1) In theoretical perspective, machine learning community has found that contrastive SSL shares many similarities with non-contrastive SSL (e.g. alignment \& uniformity~\citep{wang_understanding_2020}, Barlow Twins~\citep{zbontar_barlow_2021}, and VICReg~\citep{bardes_vicreg_2022})~\citep{balestriero_contrastive_2022,tao_exploring_2022}
(2) However, in practical applications, contrastive and non-contrastive SSL show comparable performance only in classification tasks (e.g., in Visual Representation Learning (VRL) and TR in SRL~\citep{farina_non-contrastive_2023}). In contrast, for ranking tasks (i.e., STS in SRL), contrastive SSL seems be the only effective method, significantly outperforming non-contrastive SSL~\citep{nie_inadequacy_2023,klein_scd_2022,xu_simcse_2023}. These inconsistent conclusions suggest that to make the obtained representations suitable for STS, certain unique factors must be present in the optimization objectives, which have rarely been explored in the existing literature.

In this work, we attempt to identify the key factors that enable contrastive SSL to be effective in STS. Specifically, we would like to answer two questions: (1) \textit{What commonalities enable various contrastive losses to achieve superior performance in STS?} (2) \textit{How can we make non-contrastive SSL, which is similar to contrastive SSL but ineffective in STS, effective?} We first analyze the commonalities among four effective losses~\citep{oord_representation_2018,zhang_contrastive_2022,nie_inadequacy_2023} in SRL from the perspective of gradients. From this analysis, we find that all gradients can be unified into the same paradigm, which is determined by three components: the \textbf{Gradient Dissipation}, the \textbf{Weight}, and the \textbf{Ratio}. 
By statistically analyzing the values of these three components under different representation space distributions, 
we propose three conjectures, each corresponding to the role of a component in optimizing the representation space.
Subsequently, we construct a baseline model to empirically validate our conjectures
and demonstrate the significance of these components to model performance by varying them in the baseline.

After understanding the key factors that enable contrastive losses to be effective, we are able to analyze the reasons behind the poor performance of non-contrastive SSL in STS from the perspective of three components in the paradigm. We find that these ineffective losses do not perform as well as effective ones across these components. Therefore, by adjusting these components, we manage to make them function and achieve improved model performance in STS (refer to Figure~\ref{fig:performance_0}).

Briefly, our main contributions are as follows:
\begin{itemize}
    \item We propose a unified gradient paradigm for effective losses in SRL, which is controlled by three components: the \textbf{Gradient Dissipation}, the \textbf{Weight}, and the \textbf{Ratio} (Section~\ref{sec:paradigm});
    \item We analyze the roles in optimization for each component theoretically. Further, we propose and validate the conjectures on their effective roles in performing STS tasks (Section~\ref{sec:role});
    \item With the guidance of our analysis results, we modify four optimization objectives in non-contrastive SSL to be effective in STS by adjusting the three components (Section~\ref{sec:application}).
\end{itemize}



\section{A Unified Paradigm for Gradient}
\label{sec:paradigm}

\begin{table*}[ht]\small
  \centering
    \begin{tabular}{c|l|l|l}
    \toprule
    \textbf{Objective} & \multicolumn{1}{c|}{\textbf{Gradient Dissipation}} & \multicolumn{1}{c|}{\textbf{Weight}} & \multicolumn{1}{c}{\textbf{Ratio}} \\
    \midrule
    InfoNCE~\citeyearpar{oord_representation_2018}
    &       
    $
        \begin{aligned}
        1/(1 + \frac{e^{\cos(\theta_{ii'})/\tau}}{\sum_{k\neq i}^N e^{\cos(\theta_{ik'})/\tau}})
        \end{aligned}
    $
    &       
    $\begin{aligned}
        \frac{e^{\cos(\theta_{ij'})/\tau}}{\tau\sum_{k\neq i}^N e^{\cos(\theta_{ik'})/\tau}}
    \end{aligned}$
    &  
    \multicolumn{1}{c}{1}
    \\
    ArcCon~\citeyearpar{zhang_contrastive_2022}
    &       
    $\begin{aligned}
        1/(1 + \frac{e^{\cos(\theta_{ii'} + u)/\tau}}{\sum_{k\neq i}^N e^{\cos(\theta_{ik'})/\tau}})
    \end{aligned}$
    &       
    $\begin{aligned}
        \frac{e^{\cos(\theta_{ij'})/\tau}}{\tau\sum_{k\neq i}^N e^{\cos(\theta_{ik'})/\tau}}
    \end{aligned}$
    &  
    $\begin{aligned}
        \frac{\sin(\theta_{ii'}+u)}{\sin(\theta_{ii'})}
    \end{aligned}$
    \\
    MPT~\citeyearpar{nie_inadequacy_2023}
    &       
    $\begin{aligned}
        \mathbb{I}_{\{\cos(\theta_{ii'})-\max\limits_{k\neq i}\cos(\theta_{ik'})<m\}}
    \end{aligned}$
    &       
    $\begin{aligned}
        \left\{\begin{aligned}
            &1,\quad\mathrm{else}\\
            &0,\quad j\neq\arg\min_{k\neq i}\theta_{ik'}
        \end{aligned}\right.
    \end{aligned}$
    &  
    \multicolumn{1}{c}{1}
    \\
    MET~\citeyearpar{nie_inadequacy_2023} 
    &       
    $\begin{aligned}
        \mathbb{I}_{\{\min\limits_{k\neq i}\sqrt{2-2\cos(\theta_{ik'})} -\sqrt{2-2\cos(\theta_{ii'})} < m\}}
    \end{aligned}$
    &       
    $\left\{\begin{aligned}
        &1/\sqrt{2-2\cos(\theta_{ij'})},\ \mathrm{else}\\
        &0,\quad j\neq\arg\min_{k\neq i}\theta_{ik'}
    \end{aligned}\right.$
    &  
    $\begin{aligned}
        \sqrt{\frac{1-\cos(\theta_{ij'})}{1-\cos(\theta_{ii'})}}
    \end{aligned}$
    \\
    \bottomrule
    \end{tabular}%
  \caption{Three components of four different contrastive losses. 
  We convert the cosine similarity (e.g., $h_i^\top h_j'$) and distance (e.g., $\|h_i-h_j'\|_2$) into angular form (e.g., $\cos(\theta_{ij'})$ and $\sqrt{2 - 2\cos(\theta_{ij'})}$, respectively).}
  \label{tab:effective_objectives}%
\end{table*}%



\subsection{Preliminary}
Given a batch of sentences $\{s_i\}_{i=1}^N$, we adopt a encoder $f(\cdot)$ to obtain two $l_2$-normalized embeddings (i.e., $h_i$ and $h_i'$) for two different augmented views of sentence $s_i$. Regarding $h_i$ as the anchor, the embedding from the same sentence (i.e., $h_i'$) is called the positive sample, while the embeddings from the other sentences (i.e., $h_j$ and $h_j', j\neq i$) are called negative samples. Then, what contrastive learning does is to increase the similarity (typically cosine similarity) between the anchor and its positive sample and decrease the similarity between the anchor and its negative samples.
\subsection{Gradient Analysis}
To understand the optimization mechanism rigorously, we choose four contrastive losses used in recent works and derive the gradient for them. Note that all the loss functions have been proven to be effective in SRL and can obtain competitive performance based on the model architecture of SimCSE \cite{gao_simcse_2021}.


\noindent\textbf{InfoNCE}~\citep{oord_representation_2018} is a widely used contrastive loss introduced to SRL by ConSERT \citep{yan_consert_2021} and SimCSE \citep{gao_simcse_2021}. It can be formed as
\begin{align*}
    \mathcal{L}^\mathrm{info}_i=-\log\frac{e^{h_i^\top h_i'/\tau}}{\sum_{j=1}^N e^{h_i^\top h_j'/\tau}},
\end{align*}
where $\tau$ is a temperature hyperparameter. The gradient of $\mathcal{L}^\mathrm{info}_i$ w.r.t $h_i$ is
\begin{align}
    \frac{\partial \mathcal{L}^\mathrm{info}_i}{\partial h_i}=\frac{\sum_{j\neq i}^Ne^{h_i^\top h_j'/\tau}(h_j'-h_i')}{\tau\sum_{k=1}^N e^{h_i^\top h_k'/\tau}}. 
\end{align}

\noindent\textbf{ArcCon}~\citep{zhang_contrastive_2022} improves InfoNCE by enhancing the pairwise discriminative power and it can be formed as
\begin{align*}
    \mathcal{L}^\mathrm{arc}_i
    =-\log\frac{e^{\cos(\theta_{ii'} + u)/\tau}}{e^{\cos(\theta_{ii'} + u)/\tau}+\sum_{j\neq i}^N e^{h_i^\top h_j'/\tau}},
\end{align*}
where $\theta_{ij}=\arccos(h_i^\top h_j)$ and $u$ is a hyperparameter. The gradient of $\mathcal{L}^\mathrm{arc}_i$ w.r.t $h_i$ is 
\begin{align}
    \frac{\partial \mathcal{L}^\mathrm{arc}_i}{\partial h_i}
    =&\frac
    {\sum_{j\neq i}^Ne^{h_i^\top h_j'/\tau}(h_j'-\frac{\sin(\theta_{ii'}+u)}{\sin(\theta_{ii'})}h_i')}{\tau(e^{\cos(\theta_{ii'} + u)/\tau}+\sum_{k\neq i}^N e^{h_i^\top h_k'/\tau})}.
\end{align}

\noindent\textbf{MPT} and {\bf MET}~\citep{nie_inadequacy_2023} are two loss functions in the form of triplet loss in SRL, which have the same form:
\begin{align*}
    \mathcal{L}^\mathrm{tri}_i=\max(0,d(h_i, h_i') - \max_{j\neq i} d(h_i, h_j')+m),
\end{align*}
where $m$ is a margin hyperparameter and $d: R^d \times R^d \rightarrow R$ is metric function. Specifically, the only difference among the three loss functions is $d(a,b)$, which is $-a^T b$ for MPT and
$||a-b||_2$ for MET. Similarly, we can obtain their gradient w.r.t $h_i$:
\begin{equation}\small
\begin{aligned}
    &\frac{\partial \mathcal{L}^\mathrm{mpt}_i}{\partial h_i}
    =\mathbb{I} \times
    (h_j'-h_i'),\\
    &\frac{\partial \mathcal{L}^\mathrm{met}_i}{\partial h_i}
    =\mathbb{I} \times
    \left(\frac{h_j'}{\|h_i-h_j'\|_2}-\frac{h_i'}{\|h_i-h_i'\|_2}\right),\\
\end{aligned}   
\end{equation}
where $j = \mathop{\arg\max} h_i^\top h_k' (k\neq i)$ and $\mathbb{I}$ is an indicator function equal to 1 only if $d(h_i, h_j')-d(h_i. h_i')<m$ satisfies.




Comparing the above gradient forms, we find that all gradients guide the anchor $h_i$ move towards its positive sample $h_i'$ and away from its negative samples $h_j'$. It implies that these various forms of loss functions have some commonalities in the perspective of the gradient. After reorganizing the gradient forms, we find that they can all be mapped into a \textbf{unified paradigm}:
\begin{align}\label{eq:paradigm}
    \frac{\partial \mathcal{L}_i}{\partial h_i}=\mathrm{GD}(\cdot)\sum_{j\neq i}^N \mathrm{W}(\cdot)(h_j'-\mathrm{R}(\cdot)h_i').
\end{align}
Three components control this paradigm:

\begin{itemize}
    \item $\mathrm{GD}(i, \{h_k,h_k'\}_{k=1}^N)$ is the \textbf{Gradient Dissipation} term, which overall controls the magnitude of the gradient;
    \item $\mathrm{W}(i,j,\{h_k,h_k'\}_{k=1}^N)$ is the \textbf{Weight} term, which controls the magnitude of the contribution of negative samples to the gradient;
    \item $\mathrm{R}(i, j, \{h_k,h_k'\}_{k=1}^N)$ is the \textbf{Ratio} term, which controls the magnitude of the contribution of positive samples to the gradient;
\end{itemize}

Table~\ref{tab:effective_objectives} shows the specific form of three components in each loss, which all have been converted into the angular form for subsequent analysis.


\section{Role of Each Component}
\label{sec:role}

\subsection{Theoretical Analysis}
\label{sec:theoretical}

\begin{figure}[ht]
    \centering
    \includegraphics[width=0.48\textwidth]{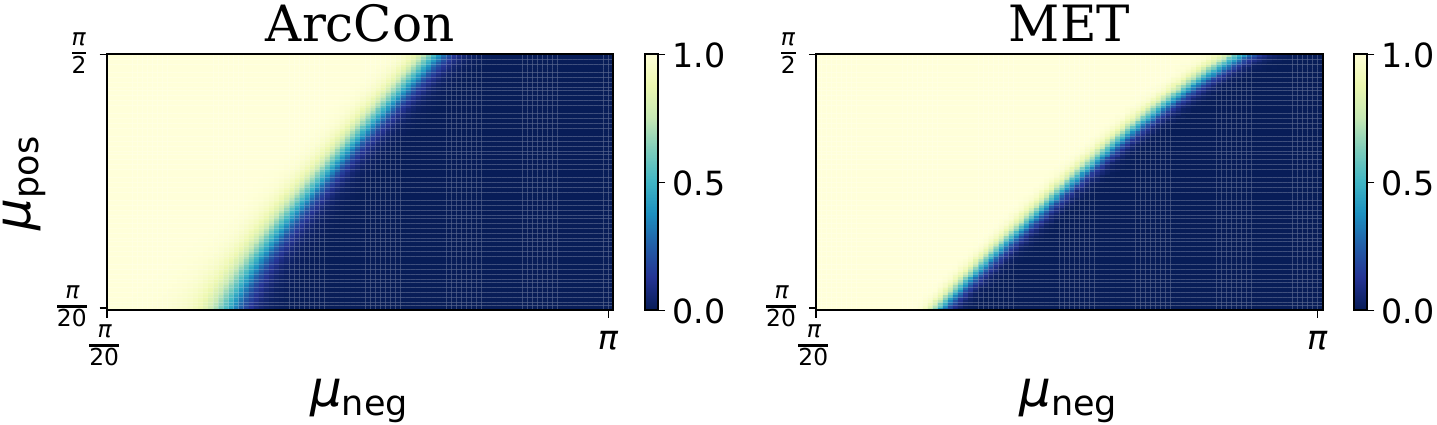}
    \caption{Average values of gradient dissipation term under different $\mu_\mathrm{pos}$-$\mu_\mathrm{neg}$ pairs for ArcCon and MET. Appendix~\ref{sec:appendix_role_illustration} shows the results for InfoNCE and MPT.}
    \label{fig:theoretical_gd_simulate}
\end{figure}

Although the above losses are unified into the same paradigm, the specific forms of the three components in different loss functions are various. Therefore, we first demonstrate that their roles are consistent despite their varied forms. To validate this, we record the trends of each component with the anchor-positive angle $\theta_{ii'}$ and anchor-negative angle $\theta_{ij'}$ and confirm whether there are consistent trends under different forms.


Specifically, we assume that $\theta_{ii'}$ and $\theta_{ij'}$ follow normal distribution $\mathcal{N}(\mu_\mathrm{pos}, \sigma_\mathrm{pos}^2)$ and $\mathcal{N}(\mu_\mathrm{neg}, \sigma_\mathrm{neg}^2)$, respectively. Based on experimental evidence, we draw $\mu_\mathrm{pos}$ from $[\frac{\pi}{20}, \frac{\pi}{2}]$, and $\mu_\mathrm{neg}$ from $[\frac{\pi}{20}, \pi]$, with $\sigma_{\mathrm{pos}}$ fixed at 0.05 and $\sigma_{\mathrm{neg}}$ fixed at 0.10. After validating the consistency among different forms, we propose intuitive conjectures regarding the roles played by each component.

\subsubsection{Gradient Dissipation}\label{sec:gradient_dissipation}


To study the role of the gradient dissipation term, we experiment with three steps: (1) For both $\mu_\mathrm{pos}$ and $\mu_\mathrm{neg}$, we divide their intervals equally into 100 parts to obtain 10,000 $\mu_\mathrm{pos}$-$\mu_\mathrm{neg}$ pairs, covering all situations of the entire optimization process; (2) For each $\mu_\mathrm{pos}$-$\mu_\mathrm{neg}$ pair, we sample 1,000 batches, each comprising 1 sampled from $\theta_{ii'}$ and 127 sampled from $\theta_{ij'}$; (3) The average value of the gradient dissipation term is calculated across these batches. The results are plotted in Figure~\ref{fig:theoretical_gd_simulate} with heatmaps. 

There are two forms of gradient dissipation: one is implemented through fraction functions (including InfoNCE and ArcCon), and the other is implemented through indicator functions (including MPT and MET). The results in Figure~\ref{fig:theoretical_gd_simulate} show that both the two forms of the gradient dissipation term exhibit a similar pattern: when $\mu_\mathrm{pos}$ and $\mu_\mathrm{neg}$ are close, the value is 1; when $\mu_\mathrm{neg}$ is larger than $\mu_\mathrm{pos}$ to some extent, the value rapidly decreases to 0. Intuitively, this term avoids a great distance gap between the anchor-positive and the anchor-negative pairs. Recall that the semantic similarity is scored on a scale from 0 to 5 in traditional STS tasks, rather than being binary classified as similar or dissimilar. Such a great gap may deteriorate the performance of sentence embeddings in STS tasks.
Therefore, we propose
\begin{conjecture}\label{conj:gd}
    The effective gradient dissipation term ensures that the distance gap between $\mu_\mathrm{neg}$ and $\mu_\mathrm{pos}$ remains smaller than the situation trained without gradient dissipation.
\end{conjecture}

\subsubsection{Weight}

\begin{figure}[ht]
    \centering
    \includegraphics[width=0.48\textwidth]{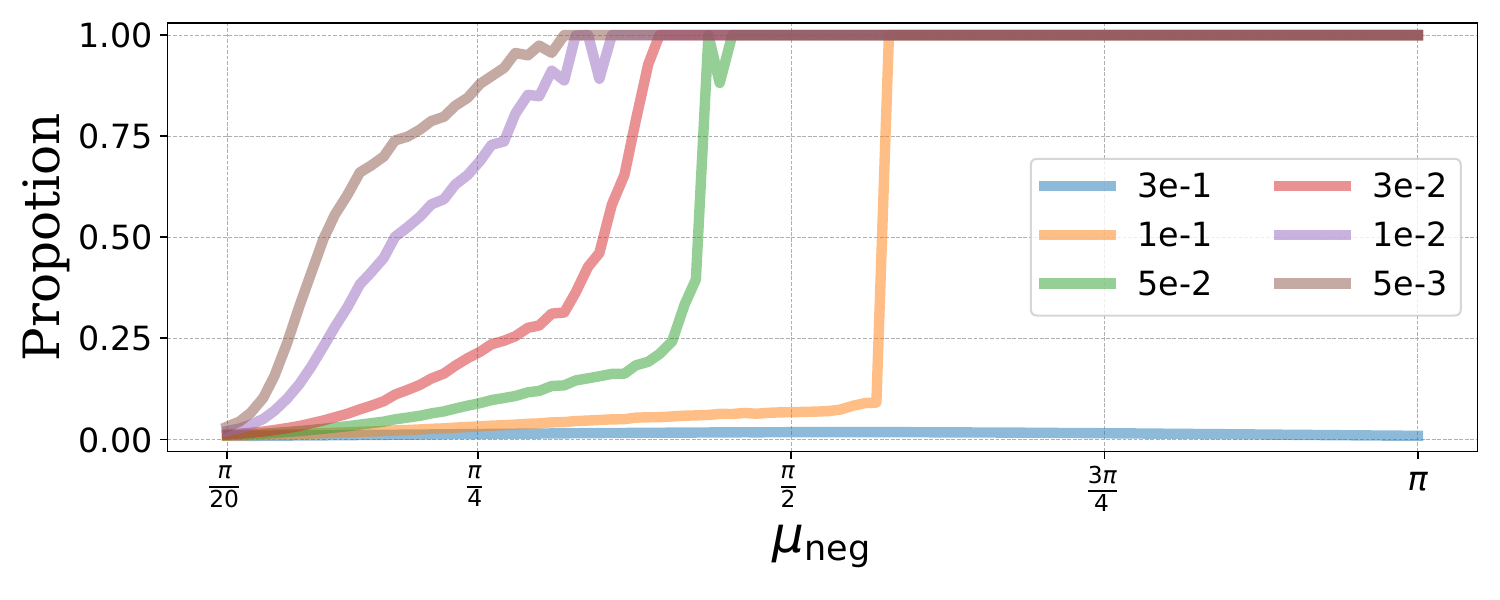}
    \caption{Variations in the average portion of the hardest negative samples in the weight across different $\mu_\mathrm{neg}$, under different temperatures $\tau$.}
    \label{fig:theoretical_w_simulate}
\end{figure}


To study the role of the weight term, we also experiment with three steps: (1) For $\mu_\mathrm{neg}$, we divide its intervals equally into 100 parts, while for $\mu_\mathrm{pos}$, we fixed its value to $\frac{\pi}{6}$. Then we can obtain 100 $\mu_\mathrm{pos}$-$\mu_\mathrm{neg}$ pairs to analyze the relative contribution among the negative samples; (2) For each $\mu_\mathrm{pos}$-$\mu_\mathrm{neg}$ pair, we sample 1000 batches, each comprising 1 sampled from $\theta_{ii'}$ and 127 sampled from $\theta_{ij'}$; (3) The average proportion of the weight for the hardest negative sample (i.e., the one with the highest cosine similarity to the anchor) is calculated.  The results calculated under different temperatures are plotted in Figure~\ref{fig:theoretical_w_simulate}. 

There are two forms of weight: one is implemented through the exponential function (including InfoNCE and ArcCon), and the other is implemented through the piecewise function (including MPT and MET). For the piecewise-form weight, the value is non-zero if and only if the negative sample is the hardest. Therefore, to prove the consistency between these two forms, it is only need to examine the proportion of the hardest negative samples in the exponential-form weight. The results show that the exponential-form weight also becomes focused solely on the hardest negative examples during the training process, 
similar to the piecewise-form weight. Therefore, we propose
\begin{conjecture}\label{conj:w}
    The effective weight term ensures that the hardest negative sample occupies a dominant position in the gradient compared to other negative samples. 
\end{conjecture}

\subsubsection{Ratio}

\begin{figure}[ht]
    \centering
    \includegraphics[width=0.48\textwidth]{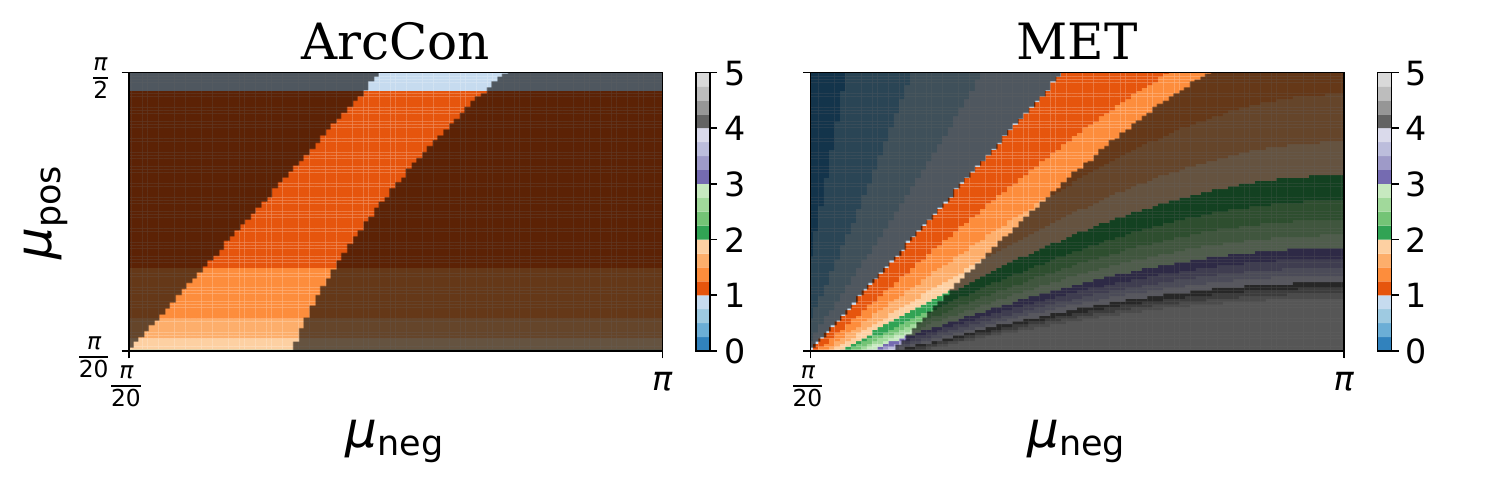}
    \caption{Average values of three dynamic ratio terms, The shaded areas indicate that these $\mu_\mathrm{pos}$-$\mu_\mathrm{neg}$ pairs do not occur in the actual optimization process, where the lower part is due to gradient dissipation, and the upper part is because there is always $\mu_\mathrm{pos} < \mu_\mathrm{neg}$. }
    \label{fig:theoretical_r_simulate}
\end{figure}


The ratio term can be categorized into two types: One is the static ratio (including InfoNCE and MPT), keeping the value of 1; The other is the dynamic ratio (including ArcCon and MET), with values dependent on $\theta_{ii'}$ and $\theta_{ij'}$. To investigate the value range of the dynamic ratio, we follow the first two steps in Section \ref{sec:gradient_dissipation} and calculate the values of different dynamic ratio terms. The results are plotted in Figure~\ref{fig:theoretical_r_simulate}. 
As shown in the figure, the dynamic ratio varies significantly. However, when considering the impact of gradient dissipation and the fact that $\mu_\mathrm{pos}$ is typically smaller than $\mu_\mathrm{neg}$, it can be found that the values of the dynamic ratio mostly fall between 1 and 2.

There exists an interesting phenomenon where sentence embeddings trained with ArcCon outperform those trained with InfoNCE, and MET does the same for MPT (Refer to Table \ref{tab:performance_1}). However, as observed in previous analyses, the gradient dissipation term and the weight term of these losses are essentially consistent. Therefore, we believe that it is precisely because ArcCon and MET can achieve larger ratios during the training process that these losses exhibit better performance, which can be explained through the following lemma.
\begin{lemma}\label{lemma:1}
    For an anchor $h_i$ and its positive sample $h_i'$ and negative sample $h_j'$, assume the angle between the plane $Oh_ih_i'$ and the plane $Oh_ih_j'$ is $\alpha$. When $h_i$ moves along the optimization direction $\lambda(rh_i'-h_j')$, $r$ must satisfy
    \begin{align*}
        r > \frac{1}{\lambda} + \frac{\sin\theta_{ij'}\cos\alpha}{\sin\theta_{ii'}} - \sqrt{\frac{1}{\lambda^2}-\frac{\sin^2\theta_{ij'}\sin^2\alpha}{\sin^2\theta_{ii'}}}.
    \end{align*}
    to ensure the distance from $h_i$ to $h_i'$ becomes closer after the optimization step.
\end{lemma}
The larger ratios in ArcCon and MET enable them to meet the condition in more situations, thereby exhibiting better performance. We propose
\begin{conjecture}\label{conj:r}
    The effective ratio term can meet the condition in Lemma~\ref{lemma:1} more frequently, and ensure that the distance from the anchor to the positive sample is closer after optimization than that before optimization.
\end{conjecture}

\subsection{Empirical Study}
\label{sec:empirical}

To validate the conjectures in Section~\ref{sec:theoretical} 
and further investigate the impact of each component on the model performance, 
this section conducts experiments based on  
\begin{align}
    \mathcal{L}_i=\mathrm{GD}(\cdot)\sum_{j\neq i}^N\mathrm{W}(\cdot)\left(h_i^\top h_j'-\mathrm{R}(\cdot)h_i^\top h_i'\right),
\end{align}
where $\mathrm{GD}(\cdot)$, $\mathrm{W}(\cdot)$, and $\mathrm{R}(\cdot)$ do not require gradient. It can be easily verified that the gradient of $\mathcal{L}_i$ w.r.t $h_i$ is the paradigm in Equation~\ref{eq:paradigm}. We select $\mathrm{GD}(\cdot)=\mathbb{I}_{\{h_i^\top h_i'-\max_{k\neq i}^Nh_i^\top h_k'<m\}}$, $\mathrm{W}(\cdot)=\frac{e^{h_i^\top h_j'/\tau}}{\sum_{k\neq i}^N e^{h_i^\top h_k'/\tau}}$, $\mathrm{R}(\cdot)=r$, and $m=0.3,\tau=0.05, r=1$ as \textbf{baseline}. We adopt BERT$_\mathrm{base}$~\citep{devlin_bert_2019} as backbone and utilize commonly used unsupervised datasets~\citep{gao_simcse_2021} as training data. 
In validating conjectures,
we divide the training data into two parts, where 90\% is used for training and 10\% is held out for statistical analysis. In investigating the impact of each component on performance, we vary each component in the baseline and train with all data, with Spearman's correlation on the STS-B~\citep{cer_semeval-2017_2017} validation set as the performance metric.



\subsubsection{Validation of Conjecture~\ref{conj:gd}}

\begin{figure}[ht]
    \centering
    \includegraphics[width=0.48\textwidth]{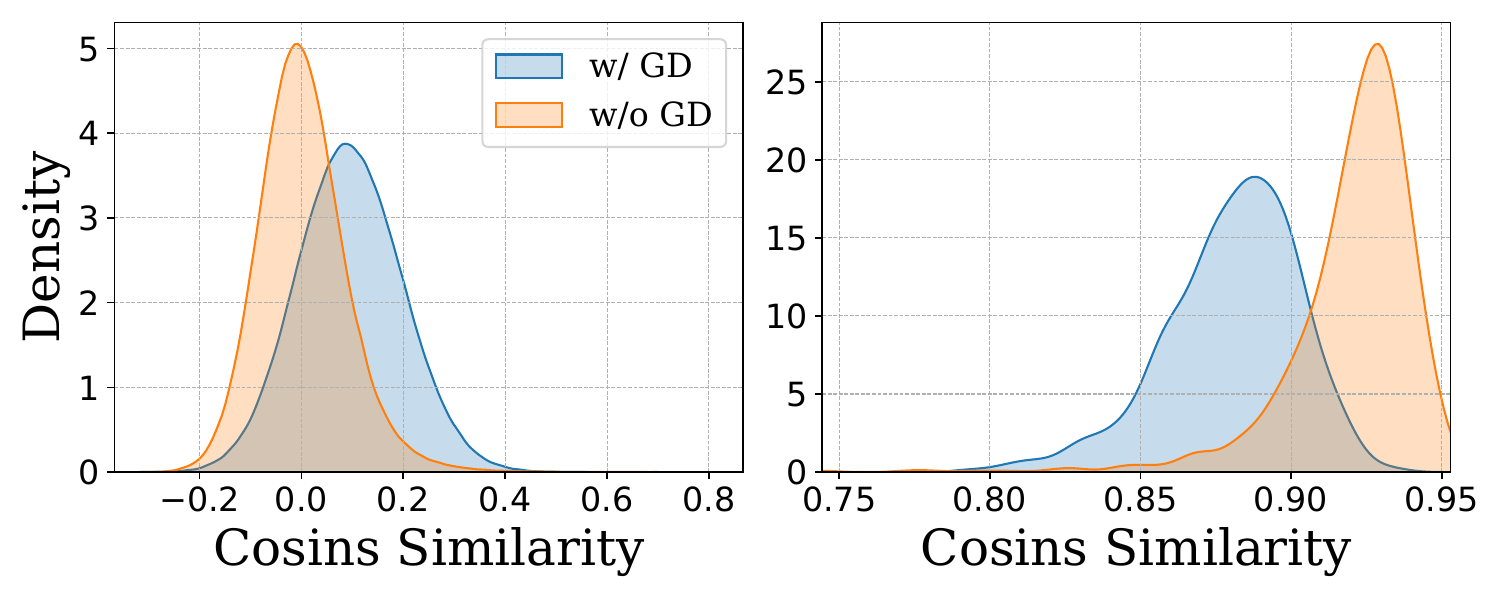}
    \caption{Distribution of cosine similarity for anchor-negative pairs (left) and anchor-positive pairs (right).}
    \label{fig:empirical_gd_cos_dist}
\end{figure}

\begin{figure*}[ht]
    \centering
    \includegraphics[width=1.0\textwidth]{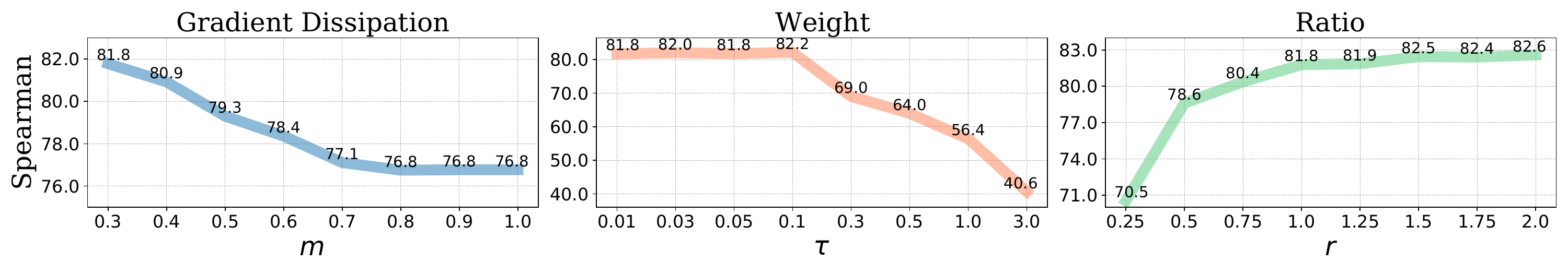}
    \caption{Spearman's correlation on the STS-B validation set when changing the three components in the baseline.}
    \label{fig:empirical_ablation}
\end{figure*}

To verify Conjecture~\ref{conj:gd}, we compare the distribution of $\theta_{ii'}$ and $\theta_{ij'}$ in sentence embeddings obtained from baseline training and from training without gradient dissipation (i.e., setting $\mathrm{GD}(\cdot)=1$). The results, presented in Figure~\ref{fig:empirical_gd_cos_dist}, indicate that the distribution of $\theta_{ii'}$ is closer to that of $\theta_{ij'}$ when gradient dissipation is applied, compared to the scenario without it, therefore validating the conjecture.

To investigate the impact of the gradient dissipation term on performance, we vary $m$ in $\mathrm{GD}(\cdot)$ from 0.3 to 1 and plot the corresponding performance changes in the first graph of Figure~\ref{fig:empirical_ablation}. An increase in $m$ implies a weakening effect, and the model performance decreasing as $m$ increases proves the importance of effective gradient dissipation term for model performance. 

\subsubsection{Validation of Conjecture~\ref{conj:w}}

\begin{figure}[ht]
    \centering
    \includegraphics[width=0.48\textwidth]{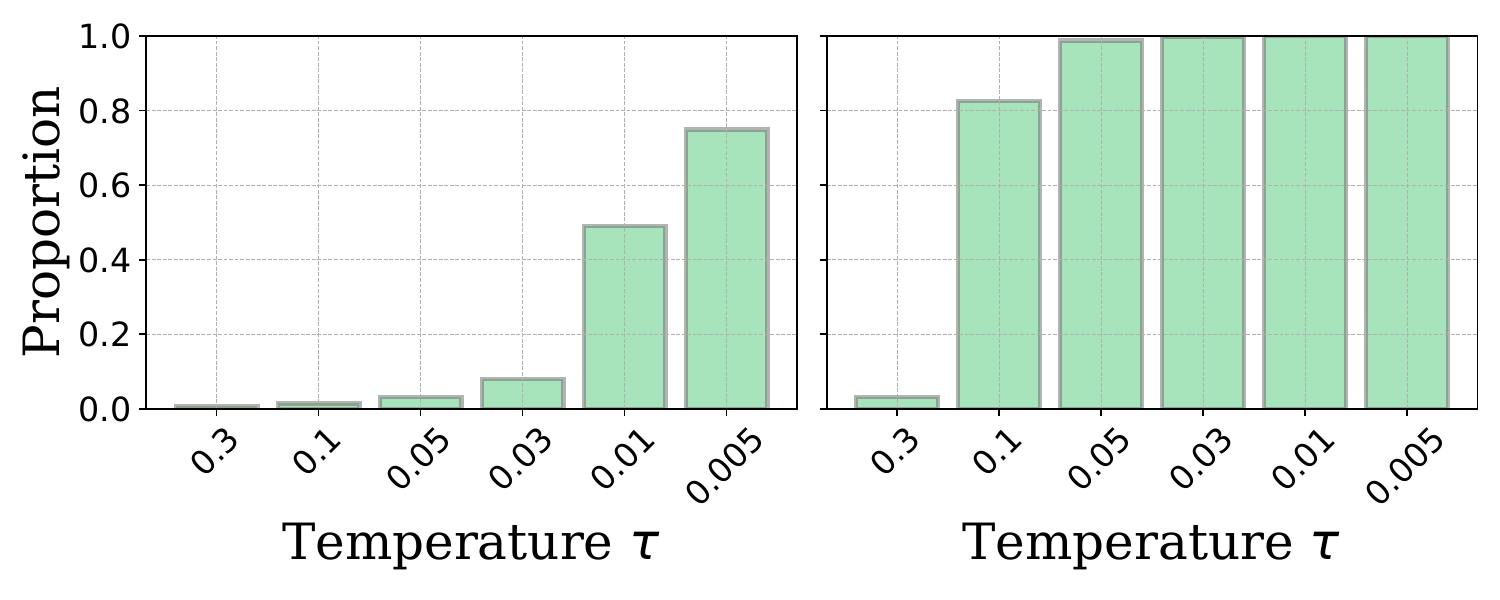}
    \caption{Average portion of the hardest
negative samples in the weight under different temperatures $\tau$, measured on the original BERT model (left) and the BERT model fine-tuned with the baseline loss (right).}
    \label{fig:empirical_w_hard_neg_percentage}
\end{figure}

To verify Conjecture~\ref{conj:w}, we calculated the average proportion of the hardest negative samples in the exponential-form weight under different temperature $\tau$, with results presented in Figure~\ref{fig:empirical_w_hard_neg_percentage}. In the original BERT, since embeddings are confined within a smaller space~\citep{li_sentence_2020}, a very small $\tau$ is required to allow the hardest negative samples to occupy a higher proportion. However, in models trained with the baseline, the spatial range of embeddings increases, enabling a commonly used setting ($\tau=0.05$) to also allow the hardest negative samples to dominate the optimization direction. These findings validate the conjecture.

To investigate the impact of the weight term on performance, we vary $\tau$ in $\mathrm{W}(\cdot)$ from 0.01 to 3.0 and plot the corresponding performance changes in the second graph of Figure~\ref{fig:empirical_ablation}. An increase in $\tau$ means the overall proportion of the hardest negative samples in the gradient decreases. When $\tau$ is increased to 0.3, the proportion begins to sharply decline (refer to Figures~\ref{fig:theoretical_w_simulate} and \ref{fig:empirical_w_hard_neg_percentage}), at which point there is also a sharp drop in model performance. This demonstrates the necessity of an effective weight term for model performance.

\begin{figure}[ht]
    \centering

    \begin{subfigure}[b]{0.48\textwidth}
        \centering
        \includegraphics[width=\textwidth]{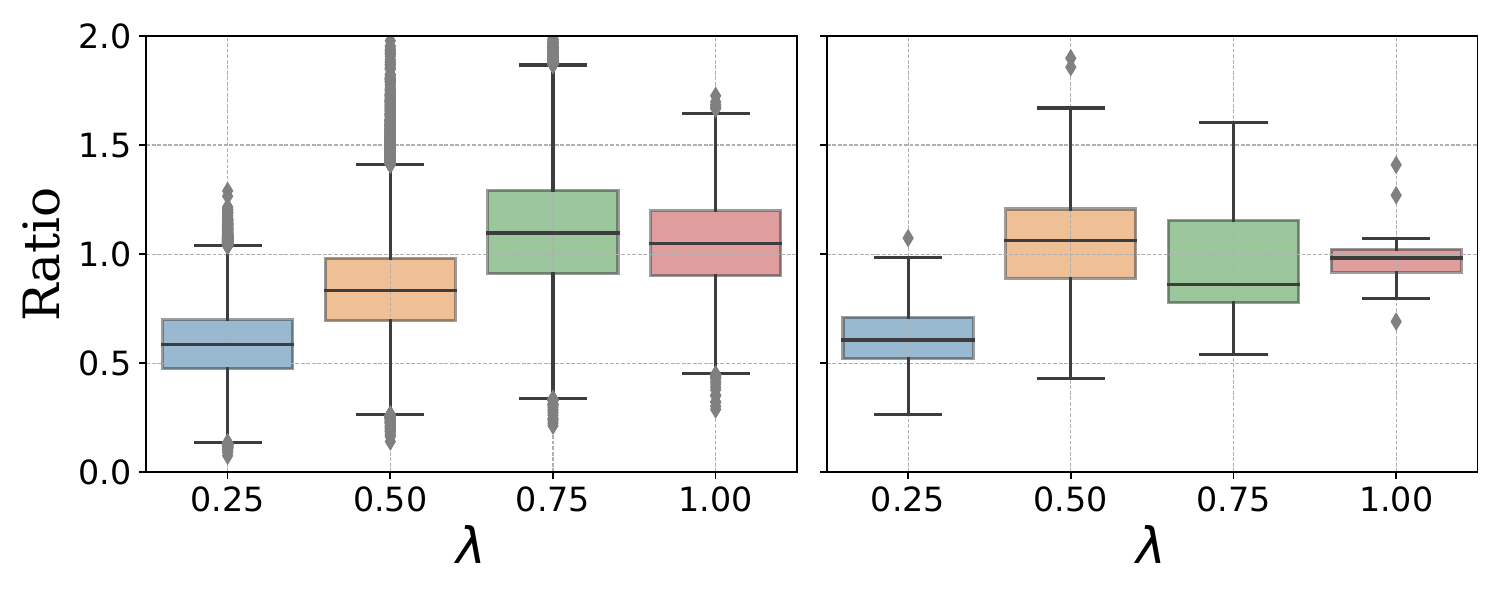}
        \caption{Distribution of the minimum ratio that satisfies the condition under different $\lambda$ for the original BERT model (left) and the BERT model fine-tuned with the baseline loss (right).}
        \label{fig:empirical_r_requirements}
    \end{subfigure}
    
    \begin{subfigure}[b]{0.235\textwidth}
        \centering
        \includegraphics[width=\textwidth]{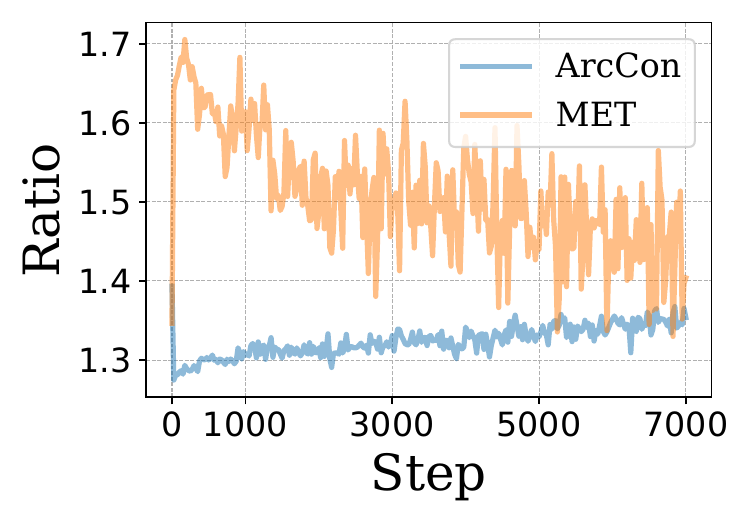}
        \caption{Variations in the dynamic ratios in ArcCon and MET across training steps.}
        \label{fig:empirical_r_dynamic_ratios}
    \end{subfigure}
    \hfill
    \begin{subfigure}[b]{0.235\textwidth}
        \centering
        \includegraphics[width=\textwidth]{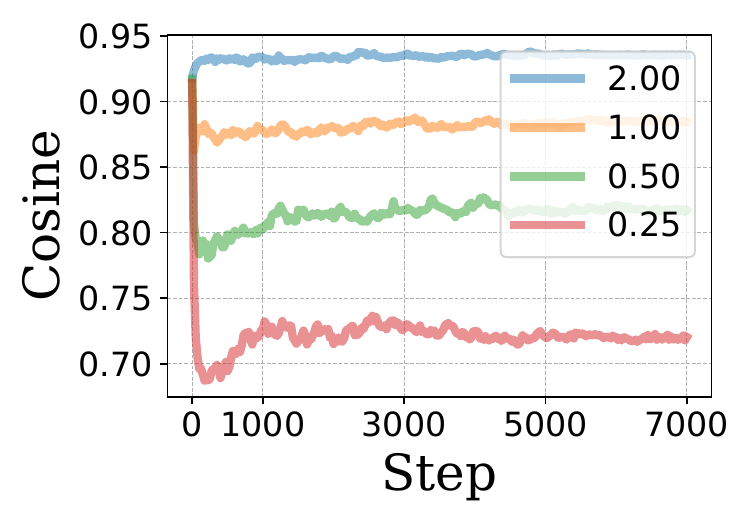}
        \caption{Variations in average cosine similarity across training steps for anchor-positive pairs.}
        \label{fig:empirical_r_cos_changes_pos}
    \end{subfigure}

    \caption{Results for validating Conjecture~\ref{conj:r}.}
    \label{fig:empirical_r}
\end{figure}

\subsubsection{Validation of Conjecture~\ref{conj:r}}
To verify Conjecture~\ref{conj:r}, 
we first record the distribution of the minimum values required to meet the conditions in Lemma~\ref{lemma:1} (presented in Figure~\ref{fig:empirical_r_requirements}). The results show that, across different models and values of $\lambda$, a ratio greater than 1 consistently meets more conditions. 
Then, we record the values of three dynamic ratios during the training process (presented in Figure~\ref{fig:empirical_r_dynamic_ratios}). The results show that ArcCon and MET maintain a ratio greater than 1 throughout the training process, thus better fulfilling the lemma's conditions. 
Finally, we record the average cosine similarity between the anchor and positive sample during the training process under different ratios (presented in Figure~\ref{fig:empirical_r_cos_changes_pos}). The results show that a larger ratio can indeed result in a closer anchor-positive distance. Together, the above results validate the conjecture.

To investigate the impact of the ratio term on performance, we vary $r$ in $\mathrm{R}(\cdot)$ from 0.25 to 2.0 and plot the corresponding performance changes in the third graph of Figure~\ref{fig:empirical_ablation}. An increase in $r$ means the conditions in Lemma~\ref{lemma:1} can be met more frequently, and the model performance improving as $r$ increases proves the significance of an effective ratio term for model performance.

\begin{table*}[ht]\small
  \centering
    \begin{tabular}{c|cccc}
    \toprule
          & $\mathcal{L}^\mathrm{a} + \nu^\mathrm{u} \mathcal{L}^\mathrm{u,MHE}$ & $\mathcal{L}^\mathrm{a} + \nu^\mathrm{u} \mathcal{L}^\mathrm{u,MHS}_i$ & $\mathcal{L}^\mathrm{B}$ & $\mathcal{L}^\mathrm{V}$ \\
    \midrule
    Gradient Dissipation 
    & 
    \multicolumn{4}{c}{\textcolor{50Red}{1}} 
    \\
    \midrule
    Weight 
    & 
    $
    \begin{aligned}
        \textcolor{50Red}{\frac{2\nu^\mathrm{u} e^{2h_i^\top h_j}}{\sum_{1\le k < l\le N}e^{2h_k^\top h_l}}}
    \end{aligned}
    $
    & 
    $
    {\scriptsize\left\{\begin{aligned}
        &\nu^\mathrm{u}/\|h_i-h_j\|_2,\ \mathrm{else}\\
        &0,j\neq\arg\max_{k\neq i}h_i^\top h_k
    \end{aligned}\right.}
    $
    &
    $
    \begin{aligned}
        \textcolor{50Red}{\frac{2\nu^\mathrm{B} h_i'^\top h_j'}{N^2}}
    \end{aligned}
    $
    & 
    $
    \begin{aligned}
        \textcolor{50Red}{\frac{4\nu^\mathrm{V,1} h_i^\top h_j}{D(N-1)^2}}
    \end{aligned}
    $ 
    \\
    \midrule
    Ratio 
    & 
    $
    \begin{aligned}
        \textcolor{50Red}{\frac{\sum_{1\le k< l\le N}e^{2h_k^\top h_l}}{\nu^\mathrm{u} N\sum_{k\neq i}^Ne^{2h_i^\top h_k}}}
    \end{aligned}
    $
    &
    $
    \begin{aligned}
        \textcolor{50Red}{\frac{2\|h_i-h_j\|_2}{\nu^\mathrm{u} N}}
    \end{aligned}
    $
    &
    $
    \begin{aligned}
        \textcolor{50Red}{\frac{N(I-(1-\nu^\mathrm{B})W_\mathrm{diag})}{\nu^\mathrm{B}\sum_{k\neq i}^N h_i'^\top h_k'}}
    \end{aligned}
    $
    &
    $
    \begin{aligned}
        \textcolor{50Red}{\frac{D(N-1)^2N^{-1}}{2\nu^\mathrm{V,1}\sum_{k\neq i}^N h_i^\top h_k}}
    \end{aligned}
    $ 
    \\
    \midrule
          & $\mathcal{L}^\mathrm{mMHE}_i$  & $\mathcal{L}^\mathrm{mMHS}_i$  & $\mathcal{L}^\mathrm{mB}_i$ & $\mathcal{L}^\mathrm{mV}_i$ \\
    \midrule
    Gradient Dissipation 
    & 
    \multicolumn{4}{c}{
    $
    \mathbb{I}_{\{h_i^\top h_i'-\max_{k\neq i}^Nh_i^\top h_k'<m\}}
    $
    } 
    \\
    \midrule
    Weight 
    &
    $
    \begin{aligned}
        \frac{e^{h_i^\top h_j/\tau}}{\tau\sum_{1\le k < l\le N}e^{\frac{h_k^\top h_l}{\tau}}}
    \end{aligned}
    $
    &
    $
    {\scriptsize\left\{\begin{aligned}
        &1/\|h_i-h_j\|_2,\ \mathrm{else}\\
        &0,j\neq\arg\max_{k\neq i}h_i^\top h_k
    \end{aligned}\right.}
    $
    &
    $
    \begin{aligned}
        \frac{e^{h_i'^\top h_j'/\tau}}{\sum_{k\neq l}^Ne^{h_k'^\top h_l'/\tau}}
    \end{aligned}
    $
    &
    $
    \begin{aligned}
        \frac{e^{h_i^\top h_j/\tau}}{\sum_{k\neq l}^Ne^{h_k^\top h_l/\tau}}
    \end{aligned}
    $
    \\
    \midrule
    Ratio 
    & 
    \multicolumn{4}{c}{$r$} 
    \\
    \bottomrule
    \end{tabular}%
  \caption{
  The three components in the optimization objectives before and after modification, with the red items indicating the parts to be modified.
  }
  \label{tab:ineffective_objectives}%
\end{table*}%

\begin{table*}[htbp]\small
  \centering
    \begin{tabular}{lcccccccc}
    \toprule
    Type  & STS12 & STS13 & STS14 & STS15 & STS16 & STS-B & SICK-R & Avg. \\
    \midrule
    \midrule
    \multicolumn{9}{c}{BERT$_\mathrm{base}$} \\
    \midrule
    $\mathcal{L}_i^\mathrm{info}$  & 68.40 & 82.41 & 74.38 & 80.91 & 78.56 & 76.85 & 72.23 & 76.25  \\
    $\mathcal{L}_i^\mathrm{arc}$   & 69.66 & 81.92 & 75.33 & 82.79 & 79.55 & 79.56 & 71.94 & 77.25  \\
    $\mathcal{L}_i^\mathrm{mpt}$   & - & - & - & - & - & - & - & 77.25  \\
    $\mathcal{L}_i^\mathrm{met}$   & - & - & - & - & - & - & - & 78.38  \\
    \midrule
    $\mathcal{L}_i^\mathrm{mMHE}$  & 71.48$_{\textcolor{70BGreen}{+18.2}}$ & 84.32$_{\textcolor{70BGreen}{+16.2}}$ & 76.54$_{\textcolor{70BGreen}{+19.6}}$ & 83.21$_{\textcolor{70BGreen}{+16.9}}$ & 80.52$_{\textcolor{70BGreen}{+12.8}}$ & 80.31$_{\textcolor{70BGreen}{+18.0}}$ & 72.41$_{\textcolor{70BGreen}{+8.6}}$ & 78.40$_{\textcolor{70BGreen}{+15.8}}$   \\
    $\mathcal{L}_i^\mathrm{mMHS}$  & 70.93$_{\textcolor{70BGreen}{+\ 7.8}}$ & 84.16$_{\textcolor{70BGreen}{+\ 6.8}}$ & 76.31$_{\textcolor{70BGreen}{+\ 8.3}}$ & 82.95$_{\textcolor{70BGreen}{+\ 4.7}}$ & 80.67$_{\textcolor{70BGreen}{+\ 4.5}}$ & 80.34$_{\textcolor{70BGreen}{+\ 6.2}}$ & 72.52$_{\textcolor{70BGreen}{+\ 0.5}}$ & 78.27$_{\textcolor{70BGreen}{+\ 5.5}}$    \\
    $\mathcal{L}_i^\mathrm{mB}$    & 71.22$_{\textcolor{70BGreen}{+15.3}}$ & 84.01$_{\textcolor{70BGreen}{+13.7}}$ & 76.49$_{\textcolor{70BGreen}{+16.4}}$ & 83.17$_{\textcolor{70BGreen}{+13.4}}$ & 80.41$_{\textcolor{70BGreen}{+10.0}}$ & 79.94$_{\textcolor{70BGreen}{+14.4}}$ & 72.45$_{\textcolor{70BGreen}{+\ 5.8}}$ & 78.24$_{\textcolor{70BGreen}{+12.7}}$   \\
    $\mathcal{L}_i^\mathrm{mV}$    &  71.42$_{\textcolor{70BGreen}{+15.8}}$ & 84.43$_{\textcolor{70BGreen}{+13.6}}$ & 76.19$_{\textcolor{70BGreen}{+16.0}}$ & 83.13$_{\textcolor{70BGreen}{+13.4}}$ & 80.80$_{\textcolor{70BGreen}{+10.2}}$ & 80.05$_{\textcolor{70BGreen}{+14.5}}$ & 72.34$_{\textcolor{70BGreen}{+\ 5.6}}$ & 78.34$_{\textcolor{70BGreen}{+12.7}}$   \\
    \midrule
    \midrule
    \multicolumn{9}{c}{RoBERTa$_\mathrm{base}$} \\
    \midrule
    $\mathcal{L}_i^\mathrm{info}$  & 70.16 & 81.77 & 73.24 & 81.36 & 80.65 & 80.22 & 68.56 & 76.57   \\
    $\mathcal{L}_i^\mathrm{arc}$   & 67.51 & 81.94 & 73.78 & 82.04 & 80.97 & 80.28 & 69.69 & 76.60  \\
    $\mathcal{L}_i^\mathrm{mpt}$   & - & - & - & - & - & - & - & 76.42  \\
    $\mathcal{L}_i^\mathrm{met}$   & - & - & - & - & - & - & - & 77.38 \\
    \midrule
    $\mathcal{L}_i^\mathrm{mMHE}$  & 69.53$_{\textcolor{70BGreen}{+10.4}}$ & 82.75$_{\textcolor{70BGreen}{+10.2}}$ & 74.35$_{\textcolor{70BGreen}{+11.4}}$ & 82.20$_{\textcolor{70BGreen}{+\ 9.9}}$ & 79.96$_{\textcolor{70BGreen}{+\ 5.2}}$ & 80.04$_{\textcolor{70BGreen}{+\ 6.1}}$ & 69.53$_{\textcolor{70BGreen}{+\ 0.0}}$ & 76.91$_{\textcolor{70BGreen}{+\ 7.6}}$  \\
    $\mathcal{L}_i^\mathrm{mMHS}$  & 68.07$_{\textcolor{70BGreen}{+\ 4.0}}$ & 82.32$_{\textcolor{70BGreen}{+\ 3.5}}$ & 73.63$_{\textcolor{70BGreen}{+\ 3.9}}$ & 82.15$_{\textcolor{70BGreen}{+\ 2.5}}$ & 79.77$_{\textcolor{70BGreen}{+\ 0.0}}$ & 79.42$_{\textcolor{45Red}{-\ 0.1}}$ & 69.25$_{\textcolor{45Red}{-\ 1.5}}$ & 76.37$_{\textcolor{70BGreen}{+\ 1.8}}$   \\
    $\mathcal{L}_i^\mathrm{mB}$    & 69.78$_{\textcolor{70BGreen}{+11.4}}$ & 82.90$_{\textcolor{70BGreen}{+\ 9.6}}$ & 74.83$_{\textcolor{70BGreen}{+11.6}}$ & 82.39$_{\textcolor{70BGreen}{+\ 9.5}}$ & 80.56$_{\textcolor{70BGreen}{+\ 5.4}}$ & 80.17$_{\textcolor{70BGreen}{+\ 5.4}}$ & 69.45$_{\textcolor{45Red}{-\ 0.7}}$ & 77.15$_{\textcolor{70BGreen}{+\ 7.5}}$   \\
    $\mathcal{L}_i^\mathrm{mV}$    & 69.06$_{\textcolor{70BGreen}{+10.9}}$ & 82.75$_{\textcolor{70BGreen}{+\ 9.6}}$ & 74.57$_{\textcolor{70BGreen}{+11.6}}$ & 82.62$_{\textcolor{70BGreen}{+\ 9.9}}$ & 79.91$_{\textcolor{70BGreen}{+\ 4.8}}$ & 79.87$_{\textcolor{70BGreen}{+\ 5.2}}$ & 69.32$_{\textcolor{45Red}{-\ 0.7}}$ & 76.87$_{\textcolor{70BGreen}{+\ 7.3}}$   \\
    \bottomrule
    \end{tabular}%
  \caption{Performance on seven \textbf{STS} tasks~\citep{conneau_senteval_2018} of the four contrastive losses, whose results are from their original paper, and the modified optimization objectives, whose results are the average value obtained from three runs. The subscript values indicate improvements compared to before modification.}
  \label{tab:performance_1}%
\end{table*}%

\section{Modification to Ineffective Losses}
\label{sec:application}

After understanding the properties that the gradients of the effective losses in SRL should possess, we can make some ineffective optimization objectives in non-contrastive SSL effective.

\noindent{\bf Alignment \& Uniformity}~\citep{wang_understanding_2020} are two metrics used to assess the performance of the representation space, yet directly using them as optimization objectives results in poor performance. Alignment can be represented as
\begin{align*}
    \mathcal{L}^\mathrm{a} = \frac{1}{N}\sum_{i=1}^N\|h_i-h_i'\|_2^2,
\end{align*}
and uniformity can be represented in many forms~\citep{liu_learning_2021}, one of which is the \textbf{Minimum Hyperspherical Energy (MHE)}
\begin{align*}
    \mathcal{L}^\mathrm{u,MHE}=\log\frac{2}{N(N-1)}\sum_{1\le k< l\le N}e^{-\|h_k-h_l\|_2^2},
\end{align*}
and another is the \textbf{Maximum Hyperspherical Separation (MHS)}
\begin{align*}
    \mathcal{L}^\mathrm{u,MHS}_i=-\min_{j\neq i}\|h_i-h_j\|_2.
\end{align*}
Another two widely used optimization objectives in non-contrastive SSL are \textbf{Barlow Twins}~\citeyearpar{zbontar_barlow_2021}
\begin{align*}
    \mathcal{L}^\mathrm{B}=\sum_{k=1}^D(C_{kk}^\mathrm{B}-1)^2+\nu^\mathrm{B}\sum_{k\neq l}^D(C_{kl}^\mathrm{B})^2,
\end{align*}
where $C^\mathrm{B}=\frac{1}{N}\sum_{i=1}^Nh_ih_i'^\top$, and {\bf VICReg}~\citeyearpar{bardes_vicreg_2022}:
\begin{align*}      
\mathcal{L}^\mathrm{V}=\mathcal{L}^\mathrm{a}
&+\nu^\mathrm{V,1}\left(v(\{h_i\}_{i=1}^N)+v(\{h_i'\}_{i=1}^N)\right) \\
&+\nu^\mathrm{V,2}\left(c(\{h_i\}_{i=1}^N)+c(\{h_i'\}_{i=1}^N)\right), \\
v(\{h_i\}_{i=1}^N&)=\frac{1}{D}\sum_{k\neq l}^D(C^\mathrm{V}(\{h_i\}_{i=1}^N)_{kl})^2,\\
c(\{h_i\}_{i=1}^N&)=\frac{1}{D}\sum_{k=1}^D\max\left(0, \gamma-\sigma(\{h_i\}_{i=1}^N)_k\right), 
\end{align*}
where $C^\mathrm{V}(\{h_i\}_{i=1}^N)=\frac{1}{N-1}\sum\limits_{i=1}^N(h_i-\bar{h}_i)(h_i-\bar{h}_i)^\top$, and $\sigma(\cdot)$ is the sample standard deviation. They are both popular in Visual Representation Learning (VRL), yet their performances are relatively poor when applied to SRL.

The gradient w.r.t $h_i$ for alignment and uniformity ($\mathcal{L}^\mathrm{a} + \nu^\mathrm{u} \mathcal{L}^\mathrm{u,MHE}$ and $\mathcal{L}^\mathrm{a} + \nu^\mathrm{u} \mathcal{L}^\mathrm{u,MHS}_i$), Barlow Twins ($\mathcal{L}^\mathrm{B}$), and VICReg ($\mathcal{L}^\mathrm{V}$) can also be mapped into the paradigm. We present the results in the upper part of Table~\ref{tab:ineffective_objectives}, where the gradient for $\mathcal{L}^\mathrm{B}$ and $\mathcal{L}^\mathrm{V}$ is derived by \citet{tao_exploring_2022}.

Components in the paradigm of ineffective losses do not perform in the same manner as those of effective losses. For gradient dissipation terms, since their values are 1 in all ineffective losses, it implies that they have no effect. For weight terms, except for MHS, other losses cannot ensure that the hardest negative samples dominate in the gradient. For ratio terms, since they are dynamic and cannot guarantee an effective value throughout the optimization process, 
it is better to adopt a static ratio. Therefore, we
adjust these terms, with the results presented in the lower part of Table~\ref{tab:ineffective_objectives}. 

We take Barlow Twins as an example to introduce the modification method. For the specific modification process of other optimization objectives, please refer to Appendix~\ref{sec:appendix_ineffective_losses_modification}. Based on gradients, the optimization objective of Barlow Twins is equivalent to 
\begin{align*}
    \mathcal{L}^\mathrm{eB}_i=-h_i^\top w_{\mathrm{p},i}^\mathrm{B}h_i'+\nu^\mathrm{B}\sum_{j\neq i}^Nh_i^\top w_{\mathrm{n},ij}^\mathrm{B}h_j,
\end{align*}
where $w_{\mathrm{p},i}^\mathrm{B}=\frac{2(I-(1-\lambda)W_\mathrm{diag})}{N}, w_{\mathrm{n},ij}^\mathrm{B}=\frac{2h_i'^\top h_j'I}{N^2}$, and both of them do not require gradient. To adjust the gradient dissipation term, we stop the gradient of anchor $h_i$ that does not meet the condition ($h_i^\top h_i' - \max_{k\neq i}^N h_i^\top h_k' < m$), which is equivalent to multiplying the loss by the indicator function
\begin{align*}
    d_i = \mathbb{I}_{\{h_i^\top h_i'-\max_{k\neq i}^Nh_i^\top h_k'<m\}}.
\end{align*}
To adjust the weight term, we first set $\nu^\mathrm{B}=1$. Then, we modify $w_{\mathrm{n},ij}^\mathrm{B}$ to an exponential form:
\begin{align*}
    w_{\mathrm{n},ij}^\mathrm{mB}=\frac{e^{h_i'^\top h_j'/\tau}}{\sum_{k\neq l}e^{h_k'^\top h_l'/\tau}}.
\end{align*}
To adjust the ratio term, we set the loss to have a static ratio $r$, by modifying $w_{\mathrm{p},i}^\mathrm{B}$ to
\begin{align*}
    w_{\mathrm{p},i}^\mathrm{mB} =\sum_{j\neq i}^N w_{\mathrm{n},ij}^\mathrm{mB} =\frac{r\sum_{j\neq i}^Ne^{h_i'^\top h_j'/\tau}}{\sum_{k\neq l}e^{h_k'^\top h_l'/\tau}},
\end{align*}
which do not require gradients. Finally, the modified Barlow Twins can be represented as
\begin{gather}
    \mathcal{L}_i^\mathrm{mB}=d_i(-h_i^\top w_{\mathrm{p},i}^\mathrm{mB}h_i'+\sum_{j\neq i}^Nh_i^\top w_{\mathrm{n},ij}^\mathrm{mB}h_j).
\end{gather}

We evaluate all the modified optimization objectives using SentEval benchmark~\citep{conneau_senteval_2018}, with the results for the Semantic Textual Similarity (STS) tasks presented in Table~\ref{tab:performance_1} and the Transfer (TR) tasks in Table~\ref{tab:performance_2}. In the STS tasks, the modified optimization objectives show a significant improvement compared to the original ones and achieved performance comparable to mainstream contrastive SSL methods. These results indicate that our modifications successfully make these non-contrastive self-supervised learning optimization objectives effective for STS tasks. Additionally, in the TR tasks, the modified optimization objectives also show overall improvement compared to the original ones, suggesting that our modifications do not introduce any additional negative impacts on the optimization objectives.

Finally, we examine the performance improvements brought by modifying each component of the optimization objectives. Specifically, we gradually modify each component of the optimization objectives in the order of the Gradient Dissipation term, the Weight term, and the Ratio term. Then, we measure the changes in the average performance on seven STS tasks, with the results shown in Table~\ref{tab:ablation}. The results indicate that as each component is modified, the performance of the optimization objectives gradually improves, validating the importance of each effective component for the optimization objectives.

\begin{table*}[htbp]\small
  \centering
    \begin{tabular}{lcccccccc}
    \toprule
    Type  & MR & CR & SUBJ & MPQA & SST2 & TREC & MRPC & Avg. \\
    \midrule
    \midrule
    \multicolumn{9}{c}{BERT$_\mathrm{base}$} \\
    \midrule
    $\mathcal{L}_i^\mathrm{info}$  & 81.18 & 86.46 & 94.45 & 88.88 & 85.50 & 89.80 & 74.43 & 85.81  \\
    $\mathcal{L}_i^\mathrm{arc}$   & 81.61 & 86.47 & 94.96 & 89.17 & 85.94 & 88.67 & 75.73 & 86.08  \\
    $\mathcal{L}_i^\mathrm{mpt}$   & - & - & - & - & - & - & - & 87.56  \\
    $\mathcal{L}_i^\mathrm{met}$   & - & - & - & - & - & - & - & 87.94  \\
    \midrule
    $\mathcal{L}_i^\mathrm{mMHE}$  & 80.42$_{\textcolor{70BGreen}{+\ 0.9}}$ & 85.31$_{\textcolor{70BGreen}{+\ 1.2}}$ & 94.10$_{\textcolor{70BGreen}{+\ 0.0}}$ & 89.10$_{\textcolor{70BGreen}{+\ 0.7}}$ & 85.03$_{\textcolor{70BGreen}{+\ 1.2}}$ & 86.73$_{\textcolor{45Red}{-\ 0.2}}$ & 74.78$_{\textcolor{70BGreen}{+\ 1.1}}$ & 85.07$_{\textcolor{70BGreen}{+\ 0.7}}$    \\
    $\mathcal{L}_i^\mathrm{mMHS}$  & 80.40$_{\textcolor{45Red}{-\ 0.2}}$ & 85.48$_{\textcolor{45Red}{-\ 0.2}}$ & 94.31$_{\textcolor{45Red}{-\ 0.1}}$ & 89.19$_{\textcolor{45Red}{-\ 0.1}}$ & 85.45$_{\textcolor{70BGreen}{+\ 0.3}}$ & 88.13$_{\textcolor{45Red}{-\ 1.6}}$ & 74.97$_{\textcolor{45Red}{-\ 1.4}}$ & 85.42$_{\textcolor{45Red}{-\ 0.5}}$     \\
    $\mathcal{L}_i^\mathrm{mB}$    & 80.69$_{\textcolor{70BGreen}{+\ 0.6}}$ & 85.31$_{\textcolor{70BGreen}{+\ 0.8}}$ & 94.35$_{\textcolor{70BGreen}{+\ 0.2}}$ & 89.38$_{\textcolor{70BGreen}{+\ 0.8}}$ & 85.34$_{\textcolor{70BGreen}{+\ 0.6}}$ & 87.53$_{\textcolor{45Red}{-\ 1.2}}$ & 74.34$_{\textcolor{70BGreen}{+\ 0.5}}$ & 85.28$_{\textcolor{70BGreen}{+\ 0.3}}$    \\
    $\mathcal{L}_i^\mathrm{mV}$    & 80.82$_{\textcolor{70BGreen}{+\ 0.8}}$ & 86.03$_{\textcolor{70BGreen}{+\ 1.9}}$ & 94.29$_{\textcolor{70BGreen}{+\ 0.1}}$ & 89.29$_{\textcolor{70BGreen}{+\ 0.7}}$ & 85.27$_{\textcolor{70BGreen}{+\ 0.5}}$ & 87.47$_{\textcolor{45Red}{-\ 1.5}}$ & 74.67$_{\textcolor{70BGreen}{+\ 0.3}}$ & 85.40$_{\textcolor{70BGreen}{+\ 0.4}}$    \\
    \midrule
    \midrule
    \multicolumn{9}{c}{RoBERTa$_\mathrm{base}$} \\
    \midrule
    $\mathcal{L}_i^\mathrm{info}$  & 81.04 & 87.74 & 93.28 & 86.94 & 86.60 & 84.60 & 73.68 & 84.84   \\
    $\mathcal{L}_i^\mathrm{arc}$   & 83.06 & 88.29 & 94.23 & 87.43 & 88.61 & 89.73 & 75.63 & 86.71  \\
    $\mathcal{L}_i^\mathrm{mpt}$   & - & - & - & - & - & - & - & 85.10  \\
    $\mathcal{L}_i^\mathrm{met}$   & - & - & - & - & - & - & - & 85.74 \\
    \midrule
    $\mathcal{L}_i^\mathrm{mMHE}$  & 82.62$_{\textcolor{70BGreen}{+\ 0.7}}$ & 87.96$_{\textcolor{70BGreen}{+\ 0.9}}$ & 93.73$_{\textcolor{45Red}{-\ 0.4}}$ & 88.13$_{\textcolor{70BGreen}{+\ 1.4}}$ & 88.19$_{\textcolor{70BGreen}{+\ 0.9}}$ & 88.53$_{\textcolor{70BGreen}{+\ 5.1}}$ & 74.78$_{\textcolor{70BGreen}{+\ 1.0}}$ & 86.28$_{\textcolor{70BGreen}{+\ 1.4}}$   \\
    $\mathcal{L}_i^\mathrm{mMHS}$  & 82.68$_{\textcolor{45Red}{-\ 0.5}}$ & 88.24$_{\textcolor{45Red}{-\ 0.6}}$ & 93.56$_{\textcolor{45Red}{-\ 0.9}}$ & 87.51$_{\textcolor{70BGreen}{+\ 0.1}}$ & 87.22$_{\textcolor{45Red}{-\ 1.1}}$ & 89.07$_{\textcolor{45Red}{+\ 0.0}}$ & 74.92$_{\textcolor{45Red}{-\ 0.7}}$ & 86.17$_{\textcolor{45Red}{-\ 0.5}}$    \\
    $\mathcal{L}_i^\mathrm{mB}$    & 82.04$_{\textcolor{70BGreen}{+\ 0.5}}$ & 87.62$_{\textcolor{70BGreen}{+\ 0.1}}$ & 93.63$_{\textcolor{45Red}{-\ 0.6}}$ & 87.83$_{\textcolor{70BGreen}{+\ 0.9}}$ & 87.75$_{\textcolor{70BGreen}{+\ 0.6}}$ & 87.73$_{\textcolor{70BGreen}{+\ 4.7}}$ & 75.28$_{\textcolor{70BGreen}{+\ 1.8}}$ & 85.98$_{\textcolor{70BGreen}{+\ 1.1}}$    \\
    $\mathcal{L}_i^\mathrm{mV}$    & 82.17$_{\textcolor{70BGreen}{+\ 0.3}}$ & 87.62$_{\textcolor{70BGreen}{+\ 0.3}}$ & 93.54$_{\textcolor{45Red}{-\ 0.7}}$ & 87.71$_{\textcolor{70BGreen}{+\ 0.8}}$ & 87.44$_{\textcolor{70BGreen}{+\ 0.2}}$ & 88.87$_{\textcolor{70BGreen}{+\ 5.1}}$ & 75.46$_{\textcolor{70BGreen}{+\ 2.0}}$ & 86.11$_{\textcolor{70BGreen}{+\ 1.1}}$   \\
    \bottomrule
    \end{tabular}%
  \caption{Performance on seven \textbf{TR} tasks~\citep{conneau_senteval_2018} of the four contrastive losses, whose results are from their original paper, and the modified optimization objectives, whose results are the average value obtained from three runs. The subscript values indicate improvements compared to before modification.}
  \label{tab:performance_2}%
\end{table*}%


\begin{table}[htbp]\small
  \centering
    \begin{tabular}{ccccc}
    \toprule
          & original & +\ GD & +\ W & +\ R \\
    \midrule
    $\mathcal{L}^\mathrm{a} + \nu^\mathrm{u} \mathcal{L}^\mathrm{u,MHE}$ & 62.62  & 64.53  & 77.39  & 78.40  \\
    $\mathcal{L}^\mathrm{a} + \nu^\mathrm{u} \mathcal{L}^\mathrm{u,MHS}$ & 72.73  & 77.40  & -*  & 78.27  \\
    $\mathcal{L}^\mathrm{B}$ & 65.60  & 67.39 & 77.84  & 78.34  \\
    $\mathcal{L}^\mathrm{V}$ & 65.53  & 68.20  & 77.90  & 78.24  \\
    \bottomrule
    \end{tabular}%
  \caption{Spearman correlation of the average performance of STS tasks when adopting BERT$_\mathrm{base}$ as the backbone, when gradually modifying each component of the optimization objectives. *: For the Weight term in MHS already satisfies the conditions, there is no need to modify its Weight term.}
  \label{tab:ablation}%
\end{table}%

\section{Related Work}
\label{sec:related}

\noindent\textbf{Contrastive Sentence Representation Learning} is initially investigated by \citet{gao_simcse_2021} and \citet{yan_consert_2021}, followed by numerous efforts~\citeyearpar{zhang_unsupervised_2022,jiang_promptbert_2022,shi_osscse_2023,li_narrowing_2024} that enhance the method's performance on STS tasks. Beyond these performance-focused studies, \citet{nie_inadequacy_2023} explores the reasons for the success of contrastive SRL, identifying the significance of gradient dissipation in optimization. The distinction of our work lies in further identifying other critical factors and  
making previously ineffective losses effective.

\noindent\textbf{Similarities between Contrastive and Non-contrastive Self-Supervised Learning} are explored in numerous works~\citeyearpar{zhang_contrastive_2022,garrido_duality_2023} in Computer Vision (CV). \citet{zhang_contrastive_2022} study the similarities between the two methods from the perspective of spectral embedding. \citet{garrido_duality_2023} point out that under certain assumptions, the two methods are algebraically equivalent. Our work makes non-contrastive SSL effective in SRL, revealing that similar parallels also exist in NLP.

\noindent\textbf{Gradients of Self-Supervised Learning} are investigated in CV by \citet{tao_exploring_2022}, which is similar to our methods.  The distinction of our work lies in that we work in the field of NLP, and that we explore the impact of different gradient components on optimizing the representation space.

\section{Conclusion}

In this paper, we propose a unified gradient paradigm for four different optimization objectives in SRL, 
which is determined by three components: the Gradient Dissipation term, the Weight term, and the Ratio term. We uncover the roles these components play in optimization and demonstrate their significance to the model performance on the STS tasks. Based on these insights, we succeed in making ineffective non-contrastive SSL optimization objectives effective in STS. Our work advances the understanding of why contrastive SSL can be effective in SRL and guides the future design of new optimization objectives.

\section*{Limitations}

First, our work currently focuses only on the impact of optimization objectives on model performance. This means that the results of this study cannot be applied to analyze the impact of model architecture on model performance. Secondly, we conduct experiments only on STS tasks in the domain of SRL, without extending to other ranking tasks or modalities. These areas are left for future work to explore. Lastly, modifying the gradient-equivalent form of optimization objectives results in significant differences in the form of optimization objectives before and after modification. To ensure consistency in form, one should make modifications based on observations and experience (see examples in Appendix~\ref{sec:appendix_ineffective_losses_modification}).

\section*{Acknowledgements}

This work was supported by the National Science and Technology Major Project under Grant 2022ZD0120200, in part by the National Natural Science Foundation of China (No. U23B2056), in part by the Fundamental Research Funds for the Central Universities, and in part by the State Key Laboratory of Software Development Environment.


\bibliography{ACL24}
\bibliographystyle{acl_natbib}

\appendix

\section{Experiment Details}
\label{sec:appendix_experiment_setup}

\subsection{Distribution of Representation Space}
\label{sec:appendix_representation_space_distribution}

\begin{figure}[ht]
    \centering
    \includegraphics[width=0.48\textwidth]{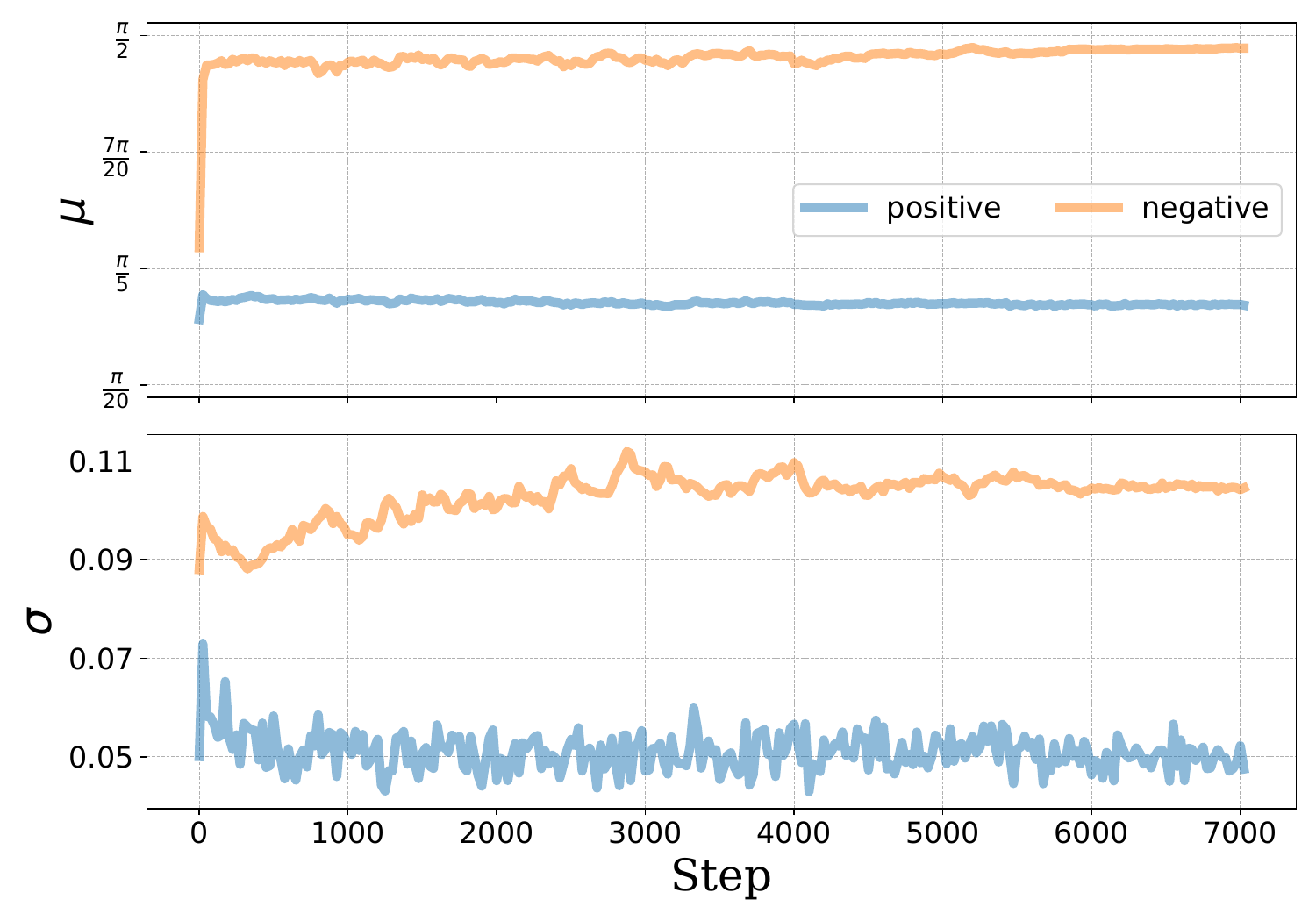}
    \caption{Variations of estimated $\mu_\mathrm{pos}$, $\mu_\mathrm{neg}$, $\sigma_\mathrm{pos}$, and $\sigma_\mathrm{neg}$ across training steps when trained with baseline.}
    \label{fig:appendix_dist_changes}
\end{figure}

In analyzing the role that each component plays in optimizing the representation space (Section~\ref{sec:theoretical}), we assume that $\theta_{ii'}$ and $\theta_{ij'}$ follow normal distributions $\mathcal{N}(\mu_\mathrm{pos},\sigma^2_\mathrm{pos})$ and $\mathcal{N}(\mu_\mathrm{neg},\sigma^2_\mathrm{neg})$, respectively. These assumptions are consistent with the phenomena observed in subsequent experiments (Section~\ref{sec:empirical}). Specifically, Figure~\ref{fig:empirical_gd_cos_dist} demonstrates that in the representation space, the distributions of $\theta_{ii'}$ and $\theta_{ij'}$ closely approximate normal distributions, and the $\theta_{ii'}$ distribution in Figure~\ref{fig:empirical_r_cos_changes_pos} is within the range of our assumptions. Furthermore, we estimate $\mu_\mathrm{pos}$, $\mu_\mathrm{neg}$, $\sigma_\mathrm{pos}$, and $\sigma_\mathrm{neg}$ during trained with the baseline. The results are presented in Figure~\ref{fig:appendix_dist_changes}. From these results, it is evident that our assumed ranges for $\mu_\mathrm{pos}$ ($[\frac{\pi}{20}, \frac{\pi}{2}]$) and $\mu_\mathrm{neg}$ ($[\frac{\pi}{20}, \pi]$) adequately cover the actual scenarios that may occur. Furthermore, the values of $\sigma_\mathrm{pos}$ and $\sigma_\mathrm{neg}$ (0.05 and 0.10, respectively) are also close to the practical observations.

\subsection{Details of the Empirical Study}

In validating the conjectures, we randomly hold out 10\% of the data for analyzing the representation space. To obtain results in Figure~\ref{fig:empirical_gd_cos_dist} and Figure~\ref{fig:empirical_r_cos_changes_pos}, we directly use the held-out data, and Figure~\ref{fig:empirical_gd_cos_dist} is plotted through Kernel Density Estimation (KDE). To obtain the results in Figure~\ref{fig:empirical_w_hard_neg_percentage}, Figure~\ref{fig:empirical_r_requirements}, and Figure~\ref{fig:empirical_r_dynamic_ratios}, we split the held-out data into batches with size $N=128$ and calculate the corresponding values within each batch. 

\subsection{Training Details}
\label{sec:appendix_training_details}

Our implementation of training the representation space is based on SimCSE~\citep{gao_simcse_2021}, which is currently widely used in research. Our experiments are conducted using Python3.9.13 and Pytorch1.12.1 on a single 32G NVIDIA V100 GPU. 

Following \citet{gao_simcse_2021}, we obtain sentence embeddings from the ``[CLS]'' token and apply an MLP layer solely during training.
The embeddings are trained with 1,000,000 sentences from Wikipedia, which are also collected by \citet{gao_simcse_2021}, for one epoch. Note that, in validating the conjectures, we randomly hold out 10\% of the data for analysis and only use 90\% for training.

\subsection{Parameter Setting}
\label{sec:appendix_parameter_setting}

For experiments in Section~\ref{sec:empirical}, all experiments are conducted with BERT$_\mathrm{base}$ as the backbone and batch size = 128, learning rate = 1e-5. 

For the modified losses in Section~\ref{sec:application}, we first perform a grid search on batch size $\in$ \{64, 128, 256, 512\}, learning rate $\in$ \{5e-6, 1e-5, 3e-5, 5e-5\} with $m=0.30$, $\tau=$ 5e-2, and $r=1$. Then, based on the best combination of batch size and learning rate from the first grid search, we perform the second grid search on 
$m\in$ \{0.27, 0.30, 0.33, 0.37\}, $\tau\in$ \{1e-2, 3e-2, 5e-2\}, and $r\in$ \{1.00, 1.25, 1.50, 1.75, 2.00\}. The results are presented in Table~\ref{tab:appendix_para_bert}
 and Table~\ref{tab:appendix_para_roberta}.

\begin{table}[ht]
  \centering
    \begin{tabular}{cccccc}
    \toprule
          & bs    & lr    & $m$     & $\tau$   & $r$ \\
    \midrule
    $\mathcal{L}^\mathrm{mMHE}_i$   & 128 & 1e-5 & 0.30 & 5e-2 & 1.75 \\
    $\mathcal{L}^\mathrm{mMHS}_i$   & 128 & 1e-5 & 0.30 & - & 1.75 \\
    $\mathcal{L}^\mathrm{mB}_i$    & 128 & 1e-5 & 0.30 & 5e-2 & 1.50 \\
    $\mathcal{L}^\mathrm{mV}_i$     & 128 & 1e-5 & 0.30 & 5e-2 & 1.50 \\
    \bottomrule
    \end{tabular}%
  \caption{The hyperparameters used to to obtain the results of modified losses in Table~\ref{tab:performance_1} when using BERT$_\mathrm{base}$~\citep{devlin_bert_2019} as the backbone.}
  \label{tab:appendix_para_bert}%
\end{table}%

\begin{table}[ht]
  \centering
    \begin{tabular}{cccccc}
    \toprule
          & bs    & lr    & $m$     & $\tau$   & $r$ \\
    \midrule
    $\mathcal{L}^\mathrm{mMHE}_i$   & 512 & 1e-5 & 0.30 & 5e-2 & 1.25 \\
    $\mathcal{L}^\mathrm{mMHS}_i$   & 512 & 1e-5 & 0.27 & - & 1.00 \\
    $\mathcal{L}^\mathrm{mB}_i$    & 512 & 1e-5 & 0.37 & 5e-2 & 1.25 \\
    $\mathcal{L}^\mathrm{mV}_i$     & 512 & 1e-5 & 0.37 & 5e-2 & 1.25 \\
    \bottomrule
    \end{tabular}%
  \caption{The hyperparameters used to to obtain the results of modified losses in Table~\ref{tab:performance_1} when using RoBERTa$_\mathrm{base}$~\cite{liu_roberta_2019} as the backbone.}
  \label{tab:appendix_para_roberta}%
\end{table}%

\subsection{Evaluation Protocol}
\label{sec:appendix_evaluation_protocol}

In Section~\ref{sec:empirical} and Section~\ref{sec:application}, we adopt a widely used evaluation protocol, SentEval toolkit~\citep{conneau_senteval_2018}, to evaluate the performance of SRL. SentEval includes two types of tasks: the Semantic Textual Similarity (STS) tasks and the Transfer tasks (TR). The STS task quantifies the semantic similarity between two sentences with a score ranging from 0 to 5 and takes Spearman’s correlation as the metric for performance. There are seven STS datasets included for evaluation: STS 2012-2016~\citeyearpar{agirre_semeval-2012_2012,agirre_sem_2013,agirre_semeval-2014_2014,agirre_semeval-2015_2015,agirre_semeval-2016_2016}, STS Benchmark~\citeyearpar{cer_semeval-2017_2017}, and SICK Relatedness~\citeyearpar{marelli_sick_2014}.  The TR task measures the performance of embeddings in the downstream classification task and takes Accuracy as the metric. There are also seven datasets included for the evaluation of TR task: MR~\citeyearpar{pang_seeing_2005}, CR~\citeyearpar{hu_mining_2004}, SUBJ~\citeyearpar{pang_sentimental_2004}, MPQA~\citeyearpar{wiebe_annotating_2005}, SST-2~\citeyearpar{socher_recursive_2013}, TREC~\citeyearpar{voorhees_building_2000}, MRPC~\citeyearpar{dolan_automatically_2005}. Note that in Section~\ref{sec:empirical}, we only use STS Benchmark validation set for evaluation, and it is conventional to use only this dataset when comparing the performance under different hyperparameters~\citep{gao_simcse_2021}.

\section{Illustrations of the Role of Each Component}
\label{sec:appendix_role_illustration}

\begin{figure}[ht]
    \centering
    \includegraphics[width=0.48\textwidth]{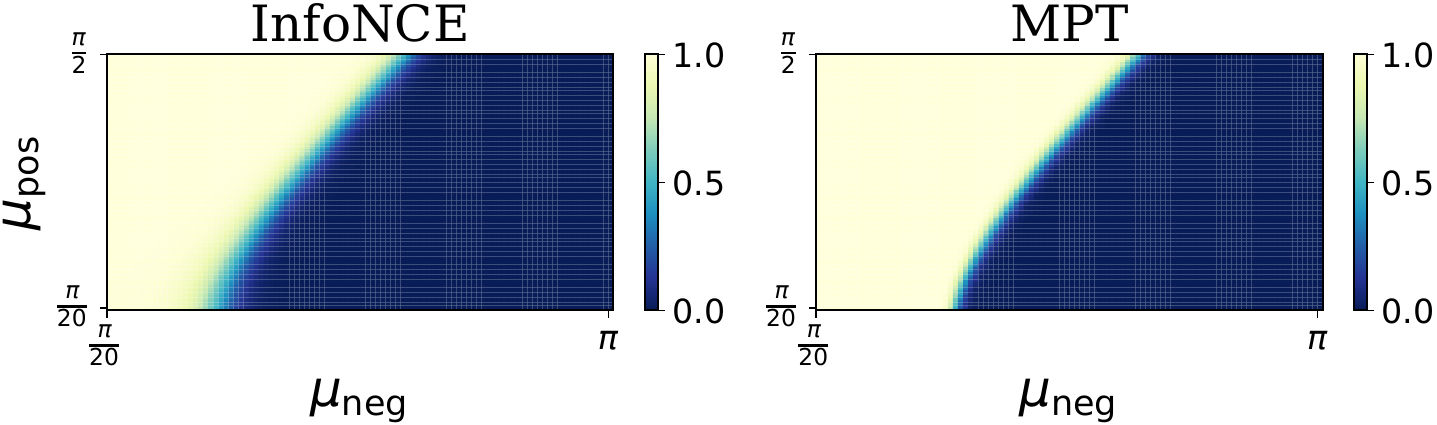}
    \caption{Average values of gradient dissipation term under different $\mu_\mathrm{pos}$-$\mu_\mathrm{neg}$ pairs for InfoNCE and MPT}
    \label{fig:appendix_gd_simulate}
\end{figure}

\begin{figure}[ht]
    \centering

    \begin{subfigure}[b]{0.48\textwidth}
        \centering
        \includegraphics[width=\textwidth]{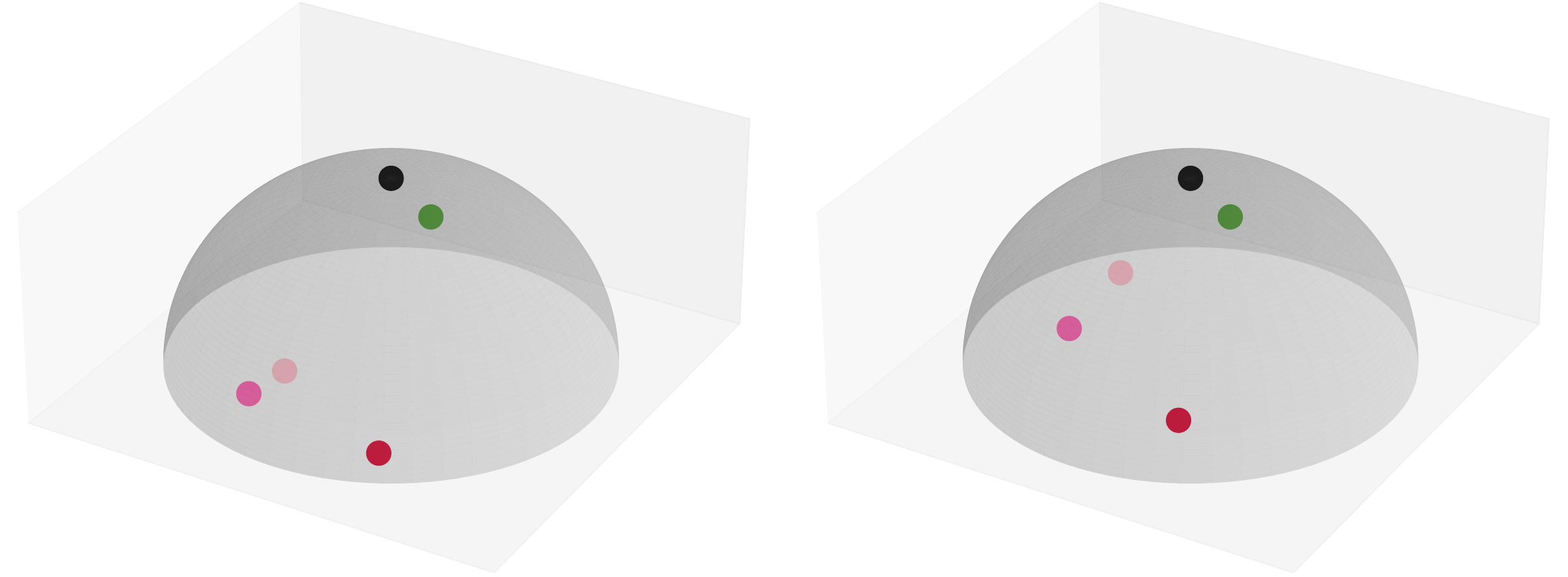}
        \caption{The gap between the anchor-positive and anchor-negative distances is larger in the representation space trained without gradient dissipation (left) compared to that with gradient dissipation (right).}
        \label{fig:appendix_gd_illustrate}
    \end{subfigure}
    
    \begin{subfigure}[b]{0.48\textwidth}
        \centering
        \includegraphics[width=\textwidth]{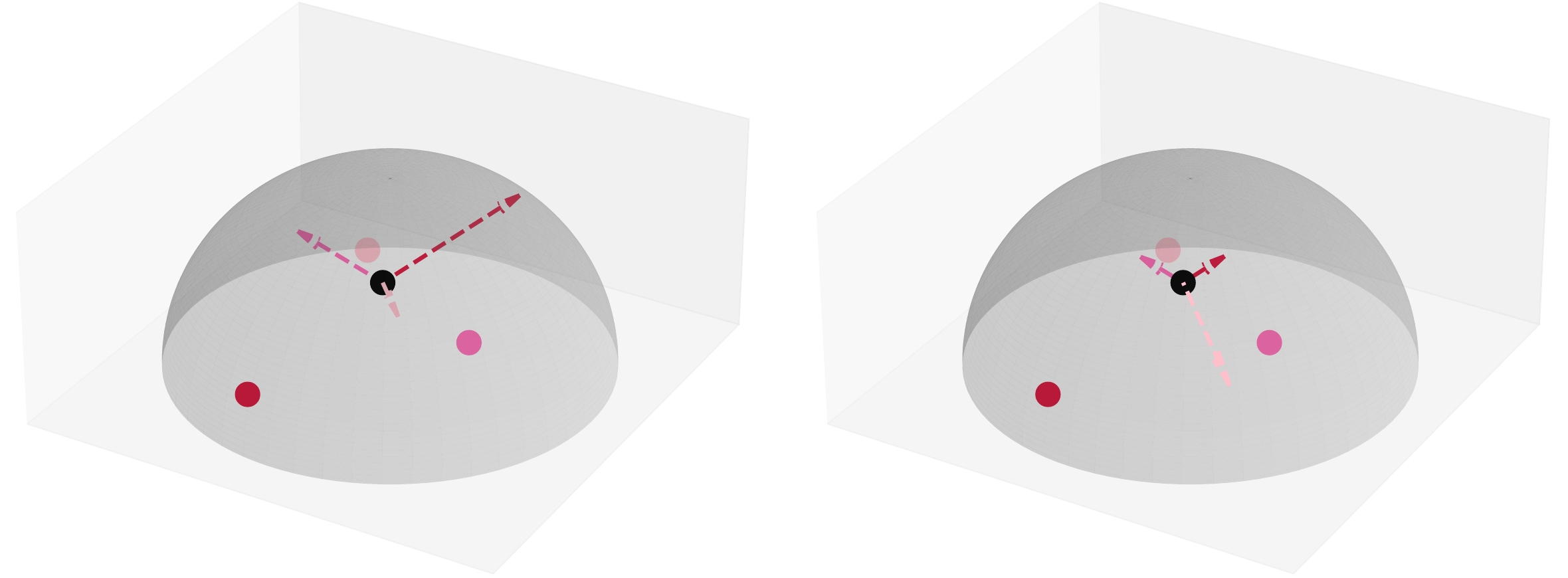}
        \caption{When not amplified by the weight term, the contribution of the hardest negative samples to the optimization direction is minimal (left). Once amplified by the weight term, the hardest negative samples dominate the optimization direction (right).}
        \label{fig:appendix_w_illustrate}
    \end{subfigure}

    \begin{subfigure}[b]{0.48\textwidth}
        \centering
        \includegraphics[width=\textwidth]{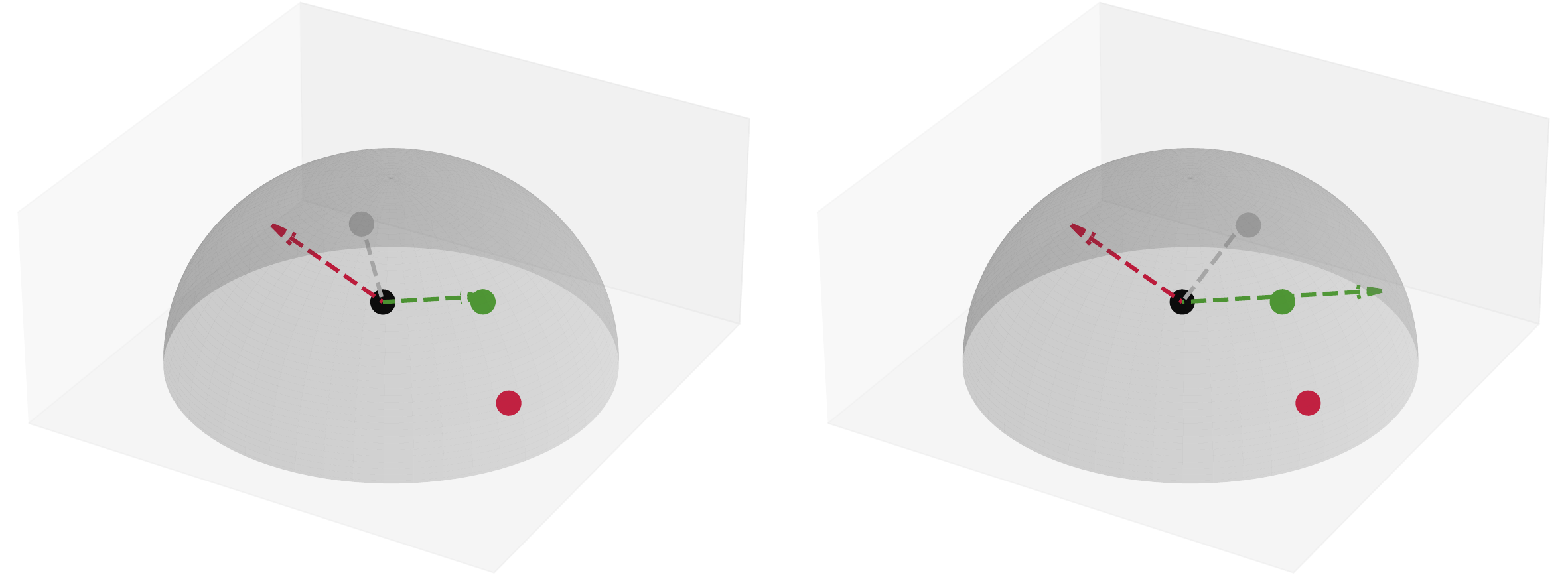}
        \caption{When the ratio term does not meet the conditions in Lemma~\ref{lemma:1}, anchor-positive distance increases after optimization. When the conditions are met, anchor-positive distance decreases after optimization.}
        \label{fig:appendix_r_illustrate}
    \end{subfigure}

    \caption{Illustrations for the role in optimization of each component.}
    \label{fig:appendix_illustration}
\end{figure}

We first present the
results of experimenting the gradient dissipation term for InfoNCE and MET (refer to Section~\ref{sec:theoretical}) in Figure~\ref{fig:appendix_gd_illustrate}.

Then, we provide three illustrations in Figure~\ref{fig:appendix_illustration} to facilitate an intuitive understanding of the roles played by different components in optimizing the representation space. In the diagram, black dots represent the anchor, gray dots represent the positions of the anchor after optimization, green dots represent positive samples, and red dots represent negative samples. Among the negative samples, the lighter the shade of red, the harder the negative sample is.

\section{Proof of Lemma~\ref{lemma:1}}
\label{sec:appendix_lemma1_proof}

\begin{figure}[ht]
    \centering
    \includegraphics[width=0.48\textwidth]{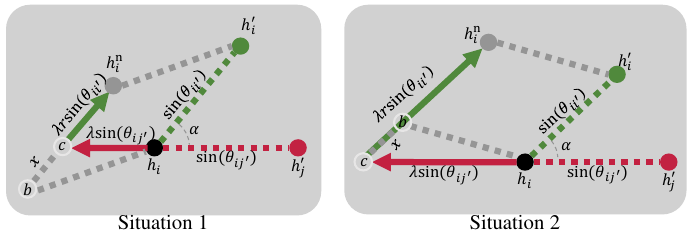}
    \caption{Illustrations of the proof to Lemma~\ref{lemma:1}}
    \label{fig:appendix_proof_lemma1}
\end{figure}

For the anchor point $h_i$, let us denote the angle between it and the positive sample $h_i'$ as $\theta_{ii'}$, and the angle between it and the negative sample $h_j'$ as $\theta_{ij'}$. Let $\alpha$ represent the dihedral angle between the planes $Oh_ih_i'$ and $Oh_ih_j'$. Under the influence of the optimization direction $\lambda(rh_i' - h_j')$, the new position of the anchor point after movement is denoted as $h_i^n$. Although all embeddings lie on a hypersphere, to simplify the proof process, we orthogonally project all embeddings onto the tangent plane at the anchor $h_i$ (as shown in Figure~\ref{fig:appendix_proof_lemma1}). In this case, the length of line $h_ih_i'$ is equal to $\sin(\theta_{ii'})$, the length of line $h_ih_j'$ is equal to $\sin(\theta_{ij'})$, and the angle $\angle h_i'h_ih_j'$ is $\alpha$. This simplification is reasonable because the optimization of the anchor $h_i$ on the hypersphere can in fact be considered as optimization on the tangent plane, which is then projected back onto the hypersphere. For the optimization direction $\lambda(rh_i' - h_j')$, we also project it onto the tangent plane, where $-\lambda h_j'$ corresponds to the line $h_ic$, and $\lambda rh_i'$ corresponds to line $ch_i^n$. 

Then, to ensure that the anchor-positive distance does not increase after optimization (i.e., $h_i^n h_i' \leq h_i h_i'$), we assume $h_i^n h_i' = h_i h_i' = \sin(\theta_{ii'})$, thereby obtaining the necessary boundary condition that satisfies the requirement. Specifically, by deriving the relationship between $r$ and the other elements at this time, we can obtain the lower bound of $r$ that satisfies the requirements. To this end, we add an auxiliary point $b$ on the line $ch_i^n$, such that the quadrilateral $bh_i^nh_i'h_i$ forms a rhombus (two possible situations when adding the auxiliary point are given in Figure~\ref{fig:appendix_proof_lemma1}), and we denote the length of the line $bc$ as $x$. By applying the cosine theorem, we can obtain the following: under Situation 1, we have 
\begin{align*}
    x_1=&-\lambda\sin(\theta_{ij'})\cos(\alpha)\\
    &+\sqrt{\sin^2(\theta_{ii'})-\lambda^2\sin^2(\theta_{ij'})\sin^2(\alpha)};
\end{align*}
under Situation 2, we have 
\begin{align*}
    x_2=&\lambda\sin(\theta_{ij'})\cos(\alpha)\\
    &-\sqrt{\sin^2(\theta_{ii'})-\lambda^2\sin^2(\theta_{ij'})\sin^2(\alpha)}.
\end{align*} 
At this point, we can express $r$ as follows:
\begin{align*}
    r
    &=\frac{\sin(\theta_{ii'})-x_1}{\lambda\sin(\theta_{ii'})}
    =\frac{\sin(\theta_{ii'})+x_2}{\lambda\sin(\theta_{ii'})}\\
    &=\frac{1}{\lambda} + \frac{\sin\theta_{ij'}\cos\alpha}{\sin\theta_{ii'}} - \sqrt{\frac{1}{\lambda^2}-\frac{\sin^2\theta_{ij'}\sin^2\alpha}{\sin^2\theta_{ii'}}},
\end{align*}
which is the lower bound of $r$ that satisfies the requirements, thus proving Lemma~\ref{lemma:1}.

\section{Gradients of Optimization Objectives}
\label{sec:appendix_objectives}

\subsection{Gradients of effective Objectives}
\label{sec:appendix_effective_losses_derivation}

In this section, we provide the derivation process of gradients in Section~\ref{sec:paradigm}. The gradient of $\mathcal{L}^\mathrm{info}_i$ w.r.t $h_i$ is
\begin{align*}
    \frac{\partial \mathcal{L}^\mathrm{info}_i}{\partial h_i}=&
    \frac{\partial}{\partial h_i}\left(-\log\frac{e^{h_i^\top h_i'/\tau}}{\sum_{j=1}^N e^{h_i^\top h_j'/\tau}}\right)\\
    =&\frac{\partial (\log\sum_{j=1}^Ne^{h_i^\top h_j'/\tau})}{\partial h_i}-\frac{\partial h_i^\top h_i'/\tau}{\partial h_i}\\
    =&\frac{\partial\sum_{j=1}^Ne^{h_i^\top h_j'/\tau} / \partial h_i}{\sum_{j=1}^Ne^{h_i^\top h_j'/\tau}}  - \frac{h_i'}{\tau}\\
    =&\frac{\sum_{j=1}^Ne^{h_i^\top h_j'/\tau} h_j'}{\tau\sum_{j=1}^Ne^{h_i^\top h_j'/\tau}} - \frac{h_i'}{\tau}\\
    =&\frac{\sum_{j\neq i}^Ne^{h_i^\top h_j'/\tau}(h_j'-h_i')}{\tau\sum_{j=1}^N e^{h_i^\top h_j'/\tau}}. 
\end{align*}
The gradient of $\mathcal{L}^\mathrm{arc}_i$ w.r.t $h_i$ is 
\begin{align*}
    \frac{\partial \mathcal{L}^\mathrm{arc}_i}{\partial h_i}
    =&\frac{\partial}{\partial h_i}\left(-\log\frac{e^{\frac{\cos(\theta_{ii'} + u)}{\tau}}}{e^{\frac{\cos(\theta_{ii'} + u)}{\tau}}+\sum\limits_{j\neq i}^N e^{\frac{h_i^\top h_j'}{\tau}}}\right)\\
    =&\frac{\partial \log (e^{\cos(\theta_{ii'} + u)/\tau}+\sum_{j\neq i}^N e^{h_i^\top h_j'/\tau})}{\partial h_i}\\
    &-\frac{\partial\cos(\theta_{ii'} + u)/\tau}{\partial h_i}\\
    =&\frac{\partial(e^{\cos(\theta_{ii'} + u)/\tau}+\sum_{j\neq i}^N e^{h_i^\top h_j'/\tau})/\partial h_i}{e^{\cos(\theta_{ii'} + u)/\tau}+\sum_{j\neq i}^N e^{h_i^\top h_j'/\tau}} \\
    &+\frac{\sin(\theta_{ii'}+u)}{\tau}\cdot\frac{\partial(\theta_{ii'} + u)}{\partial h_i} \\
    =&\frac{e^{\cos(\theta_{ii'} + u)/\tau}\cdot\frac{\sin(\theta_{ii'} + u)}{\sin(\theta_{ii'})}h_i'}{\tau(e^{\cos(\theta_{ii'} + u)/\tau}+\sum_{j\neq i}^N e^{h_i^\top h_j'/\tau})}\\
    &+\frac{\sum_{j\neq i}^Ne^{h_i^\top h_j'/\tau}h_j'}{\tau(e^{\cos(\theta_{ii'} + u)/\tau}+\sum_{j\neq i}^N e^{h_i^\top h_j'/\tau})}\\
    &-\frac{\sin(\theta_{ii'}+u)}{\tau\sin(\theta_{ii'})}h_i'\\
    =&\frac
    {\sum_{j\neq i}^Ne^{h_i^\top h_j'/\tau}(h_j'-\frac{\sin(\theta_{ii'}+u)}{\sin(\theta_{ii'})}h_i')}{\tau(e^{\cos(\theta_{ii'} + u)/\tau}+\sum_{j\neq i}^N e^{h_i^\top h_j'/\tau})}.
\end{align*}
The gradient of $\mathcal{L}^\mathrm{tri}_i$ w.r.t $h_i$ is 
\begin{align*}
    \frac{\partial \mathcal{L}^\mathrm{tri}_i}{\partial h_i}
    =&\mathbb{I}_{\{-d(h_i,h_i')+d(h_i,h_j')<m\}}\\
    &\times\left(\frac{\partial d(h_i,h_i')}{\partial h_i} - \frac{\partial d(h_i,h_j')}{\partial h_i}\right),
\end{align*}
where $j=\arg\min_{k\neq i}d(h_i,h_k')$. Therefore, The gradient of $\mathcal{L}^\mathrm{mpt}_i$ w.r.t $h_i$ is 
\begin{align*}
    \frac{\partial \mathcal{L}^\mathrm{mpt}_i}{\partial h_i}
    =\mathbb{I}_{\{h_i^\top h_i'-h_i^\top h_j'<m\}}(h_j'-h_i'),
\end{align*}
and the gradient of $\mathcal{L}^\mathrm{met}_i$ w.r.t $h_i$ is 
\begin{align*}
    \frac{\partial \mathcal{L}^\mathrm{met}_i}{\partial h_i}
    =&\mathbb{I}_{\{\|h_i-h_i'\|_2-\|h_i-h_j'\|<m\}}\\
    &\times\left(\frac{h_j'}{\|h_i-h_j'\|_2}-\frac{h_i'}{\|h_i-h_i'\|}\right).
\end{align*}

\subsection{Gradients of Ineffective Objectives}
\label{sec:appendix_ineffective_losses_derivation}

In this section, we provide the gradients of objectives in Section~\ref{sec:application}.The gradient of $\mathcal{L}^\mathrm{a}$ w.r.t $h_i$ is
\begin{align}
    \frac{\partial\mathcal{L}^\mathrm{a}}{\partial h_i}=-\frac{2}{N}h_i',
\end{align}
the gradient of $\mathcal{L}^\mathrm{u,MHE}$ w.r.t $h_i$ is
\begin{align}
    \frac{\partial\mathcal{L}^\mathrm{u,MHE}}{\partial h_i}=\frac{\sum_{j\neq i}^N2e^{2h_i^\top h_j}h_j}{\sum_{1\le k<l\le N}e^{2h_k^\top h_l}},
\end{align}
and the gradient of $\mathcal{L}^\mathrm{u,MHS}$ w.r.t $h_i$ is
\begin{align}
    \frac{\partial\mathcal{L}^\mathrm{u,MHS}}{\partial h_i}=\frac{1}{\|h_i-h_j\|_2}h_j, 
\end{align}
where $j=\arg\min_{j\neq i}\|h_i-h_j\|$.

The gradients of $\mathcal{L}^\mathrm{B}$ and $\mathcal{L}^\mathrm{V}$ w.r.t $h_i$ have been derived by \citet{tao_exploring_2022}:
\begin{align}
    \frac{\partial\mathcal{L}^\mathrm{B}}{\partial h_i}=\frac{2}{N}(-Ah_i'+\nu^\mathrm{B}\sum_{j\neq i}^N\frac{h_i'^\top h_j'}{N}h_j),
\end{align}
where $A=I-(1-\nu^\mathrm{B})C^\mathrm{B}_\mathrm{diag}$, and $C^\mathrm{B}_\mathrm{diag}$ is the diagonal matrix of $C^\mathrm{B}$, and
\begin{align}
    \frac{\partial\mathcal{L}^\mathrm{V}}{\partial h_i}=\frac{2}{N}(-h_i'+\nu^\mathrm{V}\sum_{j\neq i}^N\frac{h_i^\top h_j}{N}h_j),
\end{align}
where $\nu^\mathrm{V}=\frac{2\nu^\mathrm{V,1}N^2}{D(N-1)^2}$. Note that the original Barlow Twins adopts a batch normalization, and VICReg adopts a de-center operation, which is different from the commonly used $l_2$ normalization in SRL. But \citet{tao_exploring_2022} verify that these operations have a similar effect in training, therefore we use $l_2$ normalization for all losses for consistency.

One more thing to note is that the gradients of ineffective losses are mapped into the following form:
\begin{align}\label{eq:appendix_paradigm}
    \frac{\partial \mathcal{L}_i}{\partial h_i}=\mathrm{GD}(\cdot)\sum_{j\neq i}^N \mathrm{W}(\cdot)(h_j-\mathrm{R}(\cdot)h_i'),
\end{align}
which uses $h_j$ instead of $h_j'$ in Equation~\ref{eq:paradigm}. Since $h_j$ and $h_j'$ are mathematically equivalent with respect to $h_i$, we consider Equation~\ref{eq:paradigm} and \ref{eq:appendix_paradigm} to be consistent and only present Equation~\ref{eq:paradigm} in the main body of the paper. 

\section{Modifications to Ineffective Losses}
\label{sec:appendix_ineffective_losses_modification}

We offer two methods of modification: (1) modifying the gradient-equivalent form of the optimization objective, which is the method to modify Barlow Twins in Section~\ref{sec:application}, and (2) directly modifying the optimization objective itself.

For VICReg, we use the first method for modification as we do for Barlow Twins. Based on gradients, the optimization objective of VICReg is equivalent to 
\begin{align*}
    \mathcal{L}^\mathrm{eV}_i=-h_i^\top w_{\mathrm{p},i}^\mathrm{V}h_i'+\nu^\mathrm{V}\sum_{j\neq i}^Nh_i^\top w_{\mathrm{n},ij}^\mathrm{V}h_j,
\end{align*}
where $w_{\mathrm{p},i}^\mathrm{V}=\frac{2I}{N}, w_{\mathrm{n},ij}^\mathrm{V}=\frac{2h_i^\top h_jI}{N^2}$, and both of them do not require gradient. To adjust the gradient dissipation term, we stop the gradient of anchor $h_i$ that does not meet the condition ($h_i^\top h_i' - \max_{k\neq i}^N h_i^\top h_k' < m$), which is equivalent to multiplying the loss by the indicator function
\begin{align*}
    d_i = \mathbb{I}_{\{h_i^\top h_i'-\max_{k\neq i}^Nh_i^\top h_k'<m\}}.
\end{align*}
To adjust the weight term, we first set $\nu^\mathrm{V}=1$. Then, we modify $w_{\mathrm{n},ij}^\mathrm{V}$ to an exponential form:
\begin{align*}
    w_{\mathrm{n},ij}^\mathrm{mV}=\frac{e^{h_i^\top h_j/\tau}}{\sum_{k\neq l}e^{h_k^\top h_l/\tau}}.
\end{align*}
To adjust the ratio term, we set the loss to have a static ratio $r$, by modifying $w_{\mathrm{p},i}^\mathrm{V}$ to
\begin{align*}
    w_{\mathrm{p},i}^\mathrm{mV} =\sum_{j\neq i}^N w_{\mathrm{n},ij}^\mathrm{mV} =\frac{r\sum_{j\neq i}^Ne^{h_i^\top h_j/\tau}}{\sum_{k\neq l}e^{h_k^\top h_l/\tau}},
\end{align*}
which do not require gradients. Finally, the modified VICReg can be represented as
\begin{gather}
    \mathcal{L}_i^\mathrm{mV}=d_i(-h_i^\top w_{\mathrm{p},i}^\mathrm{mV}h_i'+\sum_{j\neq i}^Nh_i^\top w_{\mathrm{n},ij}^\mathrm{mV}h_j).
\end{gather}

For alignment and uniformity, we use the second method for modification. To adjust the gradient dissipation term, we stop the gradient of anchor $h_i$ that does not meet the condition ($h_i^\top h_i' - \max_{k\neq i}^N h_i^\top h_k' < m$) as before. As for the weight term, since the weight term of ($\mathcal{L}^\mathrm{a} + \nu^\mathrm{u} \mathcal{L}^\mathrm{u,MHS}$) does not require adjustment, we only adjust the weight term for ($\mathcal{L}^\mathrm{a} + \nu^\mathrm{u} \mathcal{L}^\mathrm{u,MHE}$). Specifically, We set $\nu^\mathrm{u}=1$ and introduce a temperature parameter $\tau$ to $\mathcal{L}^\mathrm{u,MHE}$:
\begin{align*}
    \mathcal{L}^\mathrm{u,mMHE}=\log\frac{2\sum_{1\le k< l\le N}e^{-\|h_k-h_l\|_2^2/(2\tau)}}{N(N-1)}.
\end{align*}
To adjust the ratio term, we modify both of them to have a static ratio $r$ by prepending a coefficient to the alignment:
\begin{align*}
   \mathcal{L}_i^\mathrm{a,mMHE}&=\underbrace{\frac{r\sum_{j\neq i}^Ne^{h_i^\top h_j/\tau}}{2\tau \sum_{1\le k<l\le N}e^{h_k^\top h_l/\tau}}}_\mathrm{no\ gradient}
   \|h_i-h_i'\|_2^2,\\
   \mathcal{L}_i^\mathrm{a,mMHS}&=\underbrace{\frac{r}{2\max_{j\neq i}\|h_i-h_j\|_2}}_\mathrm{no\ gradient}\|h_i-h_i'\|_2^2.
\end{align*}
Finally, the modified losses can be represented as
\begin{align}
    \mathcal{L}_i^\mathrm{mMHE}&=d_i(\mathcal{L}_i^\mathrm{a,mMHE}+\mathcal{L}^\mathrm{u,mMHE}),\\
    \mathcal{L}_i^\mathrm{mMHS}&=d_i(\mathcal{L}_i^\mathrm{a,mMHS}+\mathcal{L}_i^\mathrm{u,MHS}).
\end{align}

In order to verify the correctness of the modification, we present the gradients of all the modified losses here. The gradient of $\mathcal{L}_i^\mathrm{mMHE}$ w.r.t $h_i$ is
\begin{align}
    \frac{\mathcal{L}_i^\mathrm{mMHE}}{\partial h_i}=d_i\frac{\sum_{j\neq i}^Ne^{h_i^\top h_j/\tau}(h_j-rh_i')}{\tau\sum_{1\le k<l\le N}e^{h_k^\top h_l/\tau}}.
\end{align}
\begin{align}
    \frac{\mathcal{L}_i^\mathrm{mMHE}}{\partial h_i}&=d_i\frac{(h_j-rh_i')}{\|h_i-h_j\|_2}, \notag\\
    j&=\arg\min_{k\neq i}\|h_i-h_k\|_2
\end{align}
\begin{align}
    \frac{\mathcal{L}_i^\mathrm{mB}}{\partial h_i}=d_i\frac{\sum_{j\neq i}^Ne^{h_i'^\top h_j'/\tau}(h_j-rh_i')}{\sum_{k\neq l}^Ne^{h_k'^\top h_l'/\tau}}.
\end{align}
\begin{align}
    \frac{\mathcal{L}_i^\mathrm{mV}}{\partial h_i}=d_i\frac{\sum_{j\neq i}^Ne^{h_i^\top h_j/\tau}(h_j-rh_i')}{\sum_{k\neq l}^Ne^{h_k^\top h_l/\tau}}.
\end{align}


\end{document}